\documentclass[sigconf]{acmart}

\usepackage{balance}
  
\usepackage{booktabs} 
\usepackage{bm}
\usepackage{algorithm,algorithmic} 
 
\usepackage{multirow}
\usepackage{float}
\usepackage{subfigure}

\usepackage{booktabs}
\usepackage{threeparttable}

\def\BibTeX{{\rm B\kern-.05em{\sc i\kern-.025em b}\kern-.08emT\kern-.1667em\lower.7ex\hbox{E}\kern-.125emX}}
    
 \setcopyright{acmcopyright}

\begin{document}
\fancyhead{}
\title{Regularized Adversarial Sampling and Deep Time-aware Attention for Click-Through Rate Prediction}


\author{Yikai Wang$^1${\footnotemark[1]} \quad Liang Zhang$^2${\footnotemark[1]} \quad Quanyu Dai$^3$ \quad Fuchun Sun$^1${\footnotemark[6]} \quad Bo Zhang$^2$ \quad \\Yang He$^2$ \quad Weipeng Yan$^2$ \quad Yongjun Bao$^2$}

\affiliation{%
  \institution{
    $^1$ Department of Computer Science and Technology, Tsinghua University \quad\\Beijing National Research Center for Information Science and Technology (BNRist)\\
    $^2$ JD.COM \quad
    $^3$ The Hong Kong Polytechnic University} 
}
\email{{wangyk17@mails., fcsun@}tsinghua.edu.cn, csqydai@comp.polyu.edu.hk} \email{{zhangliang16, zhangbo35, landy, Paul.yan, baoyongjun}@jd.com}

%
\renewcommand{\shortauthors}{Yikai Wang and Liang Zhang, et al.}

\begin{abstract}
Improving the performance of click-through rate (CTR) prediction remains one of the core tasks in online advertising systems. With the rise of deep learning, CTR prediction models with deep networks remarkably enhance model capacities. In deep CTR models, exploiting users' historical data is essential for learning users' behaviors and interests. As existing CTR prediction works neglect the importance of the temporal signals when embed users' historical clicking records, we propose a time-aware attention model which explicitly uses absolute temporal signals for expressing the users' periodic behaviors and relative temporal signals for expressing the temporal relation between items. Besides, we propose a regularized adversarial sampling strategy for negative sampling which eases the classification imbalance of CTR data and can make use of the strong guidance provided by the observed negative CTR samples. The adversarial sampling strategy significantly improves the training efficiency, and can be co-trained with the time-aware attention model seamlessly. Experiments are conducted on real-world CTR datasets from both in-station and out-station advertising places.
\end{abstract}


\settopmatter{printacmref=false, printfolios=false}
%
%


\copyrightyear{2019}
\acmYear{2019}
\acmConference[CIKM '19]{The 28th ACM International Conference on Information and Knowledge Management}{November 3--7, 2019}{Beijing, China}
\acmPrice{15.00}
\acmDOI{10.1145/3357384.3357936}
\acmISBN{978-1-4503-6976-3/19/11}

%
\maketitle

{\fontsize{8pt}{8pt} \selectfont
\textbf{ACM Reference Format:}\\
Yikai Wang, Liang Zhang, Quanyu Dai, Fuchun Sun, Bo Zhang, Yang He, Weipeng Yan, and Yongjun Bao. 2019. Regularized Adversarial Sampling and Deep Time-aware Attention for Click-Through Rate Prediction. In \textit{The 28th ACM International Conference on Information and Knowledge Management (CIKM '19), November 3--7, 2019, Beijing, China.} ACM, New York, NY, USA, 10 pages. \url{https://doi.org/10.1145/3357384.3357936}}

\section{Introduction}
Online advertising is widely used for delivering promotional products to users, due to its low ads displaying cost, easy customization, easy deployment, large coverage and fast delivery speed. In online advertising, cost per click (CPC) is one of the dominant methods, in which the advertisers pay for each click on their ads. Click-though rate (CTR) prediction, with the objective to estimate the probability of users' clicking behaviors, can directly influence the performance of both bidding and ranking in CPC advertising systems.

Improving users' CTR prediction performance in online advertising remains a hot research topic. In the early stage, FM\cite{DBLP:conf/icdm/Rendle10} uses cross terms of user features and item features aiming to capture their combination relations. Recent years, inspired by the powerful capability of deep learning, deep CTR models like Deep Crossing\cite{DBLP:conf/kdd/WangFFW17}, Wide\&Deep\cite{DBLP:conf/recsys/Cheng0HSCAACCIA16}, DeepFM\cite{DBLP:conf/ijcai/GuoTYLH17} extend early works by replacing the transformation functions with complex networks, which enhance the model capacities. In these works, users' historical behaviors are integrally converted to low-dimensional embeddings without exploration of each individual historical item. Thus these models are limited to represent users' rich historical behaviors.

To further improve the performance of CTR prediction, a crucial part is to learn users' preferences on the basis of their past interactions with items, which are detailedly recorded over time. To do this, current works mainly use attention-based\cite{DBLP:journals/corr/BahdanauCB14, DBLP:conf/nips/SutskeverVL14, DBLP:conf/emnlp/LuongPM15} models to exploit users' historical clicking records. DIN\cite{DBLP:conf/kdd/ZhouZSFZMYJLG18} and DIEN\cite{DBLP:conf/aaai/ZhouMFPBZZG19} capture users' relative interests by exploiting users' historical clicking items using the attention mechanism.

\renewcommand*{\thefootnote}{\fnsymbol{footnote}}
\footnotetext[1]{Both authors contribute equally to this research.}
\footnotetext[6]{Fuchun Sun (fcsun@tsinghua.edu.cn) is the corresponding author.}
\renewcommand*{\thefootnote}{\arabic{footnote}}

However, these attention-based models do not explicitly use the clicking temporal signals of users' historical data. The temporal signals, on the one hand, can reflect the users' periodic behavior trends. For example, a user is likely to be more active after the payday of each month, and tends to buy various seasonal clothes in various months of each year. On the other hand, temporal signals can express the extent of the influence of each historical item on the target recommended item. Users' behaviors present rich changes with time, as past interests may fade away and new interests may emerge. In order to take advantage of such temporal information in users' historical clicking records, we propose a time-aware attention model for CTR prediction. The time-aware attention model contains absolute temporal signals for periodicity representation and relative temporal signals for temporal relation representation. Comparison results show that in CTR prediction tasks, the proposed time-aware attention model outperforms the current works by a large margin.

CTR prediction essentially aims to classify between the minority positive samples and the majority negative samples, and suffers from a serious data imbalance problem (such as 1:100). Adopting random sampling or down sampling on the negative samples can alleviate the classification imbalance to some extent\cite{DBLP:conf/kdd/HePJXLXSAHBC14}. However, these sampling strategies may miss informative negative samples, which are overwhelmed by the huge amounts of the total negatives. GAN-based\cite{DBLP:conf/nips/GoodfellowPMXWOCB14} sampling, which is a state-of-the-art method in recommender systems, improves data efficiency by seeking competitive nonpositive samples for each positive sample to promote adversarial training, where the nonpositive samples stand for arbitrary combinations of users and items that are not positive and are often non-interactive. IRGAN\cite{DBLP:conf/sigir/WangYZGXWZZ17} and AdvIR\cite{DBLP:conf/www/ParkC19} both apply adversarial sampling models in the information retrieval area, and the authors demonstrate several applications containing item recommendation. \cite{DBLP:conf/kdd/WangYHLWH18} describes recommendation-specific adversarial sampling method in detail and acquires effective performance. A same weakness of the current adversarial sampling methods in recommender systems is that, they can only select the nonpositive samples to train against the positive samples, thus these GAN sampling works are not applicable to CTR prediction tasks where observed negative samples should be considered.

\begin{figure}[ht]
\centering
\includegraphics[scale=0.43]{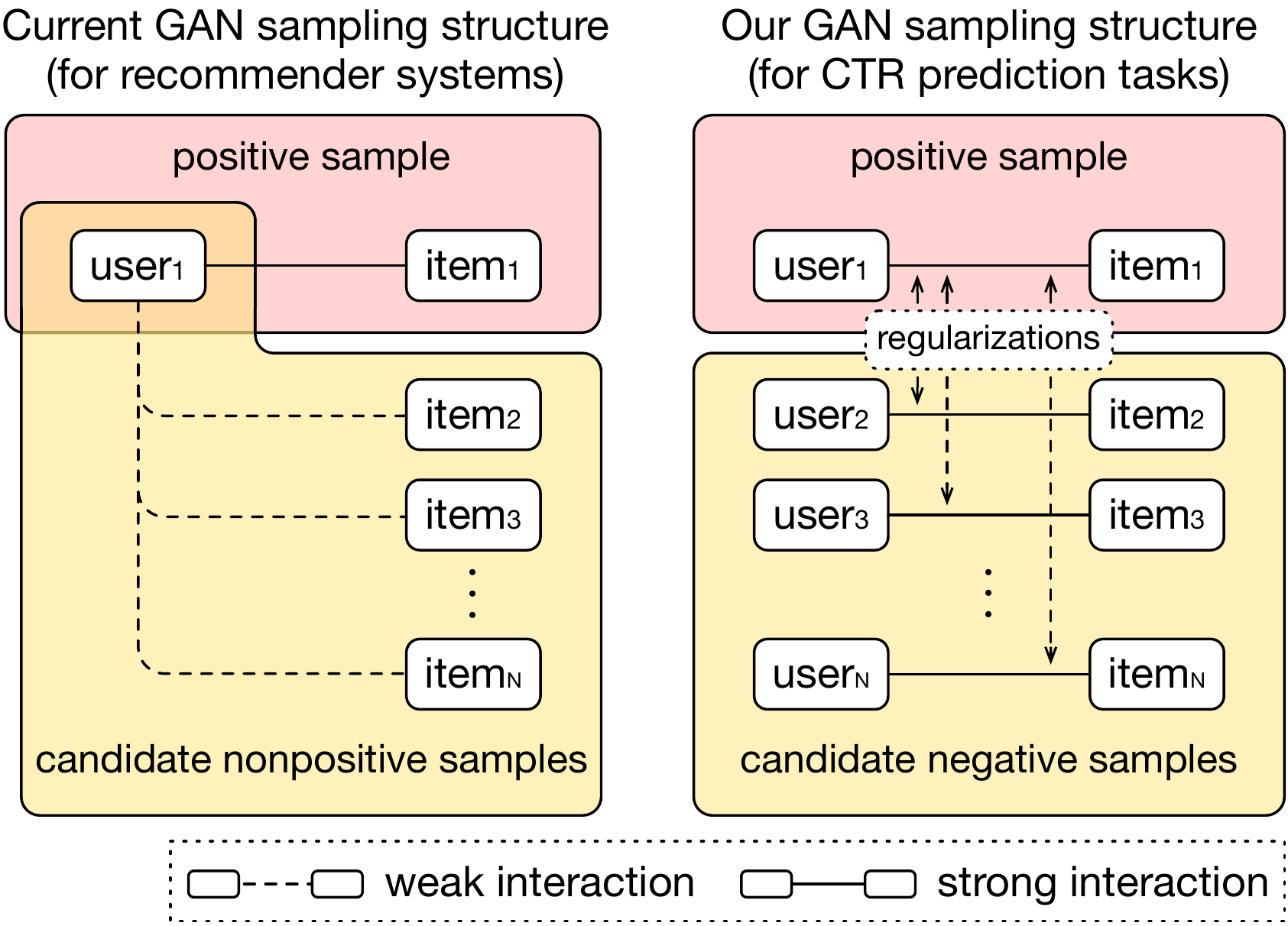}
\caption{GAN-based sampling structures comparison}
\label{structure_comparison}
\end{figure}

In CTR prediction tasks, each item is carefully selected and exposed to a user by the advertisement. Whether the exposed item acquires a click or not, it has already practically interacted with the user, and thus can provide us with strong user-item information. Recommended items that fail to be clicked, are treated as items that the user dislike, and these user-item pairs are labeled as negative samples. These observed negative samples are more negative compared with previously mentioned nonpositive samples, and thus can provide more guidance for pairwise learning. In order to utilize such guidance, we propose a regularized adversarial sampling model, which contains a distance-based discriminator and a probability-based generator. We reframe the adversarial sampling process in view of imbalanced classification and indicate that the selected negative sample needs to be competitive among all the negatives as well as correlative to the given positive sample. Therefore in the discriminator, we design a feedback that contains not only the sample score but also a regularization term. The regularization is a weighted Euclidean distance between the embeddings of the positive and negative samples calculated in the discriminator. The designed feedback will help the generator select proper negative samples that can promote the adversarial training. The difference between the adversarial sample structures of the current works for common recommender systems and ours specifically for CTR prediction tasks is illustrated in Figure \ref{structure_comparison}.

Our regularized adversarial sampling strategy is specifically designed for CTR prediction, and can be seamlessly integrated with the proposed time-aware attention model. Besides, we provide a CTR calibration method to further deal with an over-estimation issue of the absolute CTR values in the negative sampling process.

The contributions of our works are summarized as follows:
\leftmargini=7mm
\begin{itemize}
\setlength\itemsep{0em}

\item We propose a time-aware attention model for CTR prediction tasks, which can represent the users’ periodic behaviors and the
temporal relations between items, by taking absolute and relative temporal signals into consideration.
\item We design a regularized adversarial sampling strategy for CTR prediction, which can use the strong information of the observed negative samples. The adversarial sampling model can be co-trained with the time-aware attention model.
\item We conduct experiments on real-world CTR data from both in-station and out-station advertising places, and acquire comparison relative CTR results with recent state-of-the-art works. We obtain accurate absolute CTR results with the help of CTR calibration. We also provide an in-depth model exploration and sensitivity analysis of hyperparameters.
\end{itemize}

The rest of the paper is organized as follows. We propose the time-aware attention model in Section \ref{sec2} and the regularized adversarial sampling strategy in Section \ref{sec3}. We provide experimental details, comparison results and in-depth analysis in Section \ref{sec4}. We summarize the related works in Section \ref{sec5}. Finally we conclude our works in Section \ref{sec6}.

\section{Time-aware Attention}
\label{sec2}
\subsection{Data Description}
\label{sec21}
Historical user-item interactions are largely recorded and play a pivotal role in many real-world CTR prediction tasks. In our model, a single sample is composed of a pair of a user and a target item (recommended item), where the user's information is represented by a series of user's recent $L$ clicking items, and the target item is selected and exposed to the user by the advertisement. Each item has three parts of features, described as follows:

\leftmargini=7mm
\begin{itemize}
\setlength\itemsep{0em}
\item Raw features embedding, denoted as $\bm{e}^r$, has 50 dimensions and is obtained by a pre-training process on the displayed image of the item using Telepath\cite{DBLP:conf/aaai/WangXWLHHY18}. Telepath acts as a good feature extractor to our model as it extracts the key features with deep CNNs\cite{DBLP:conf/nips/KrizhevskySH12} by considering relations between the users’ historical records and the target item. 
\item Interaction time, denoted as $t$, refers to the clicking time for historical items or the exposing time for the target item, containing an absolute time $t^a$ in seconds, a month index $t^m$, a week index $t^w$, a day index $t^d$ and an hour index $t^h$.
\item Category index (cid3), denoted as $c$, is a single scalar value representing the category that the item belongs to.
\end{itemize}

In summary, each sample in our CTR data contains $L+1$ items in total, including $L$ historical clicking items with features $(\bm{e}_l^r,t_l,c_l),l=1\cdots L$, and a target item with features $(\bm{e}_0^r,t_0,c_0)$.

Besides, each sample has several auxiliary features that represent the attributes of the user such as the gender, age, etc.

\subsection{Time-aware Embedding Structure}
\label{sec:att}
\begin{figure*}[ht]
\centering
\includegraphics[scale=0.45]{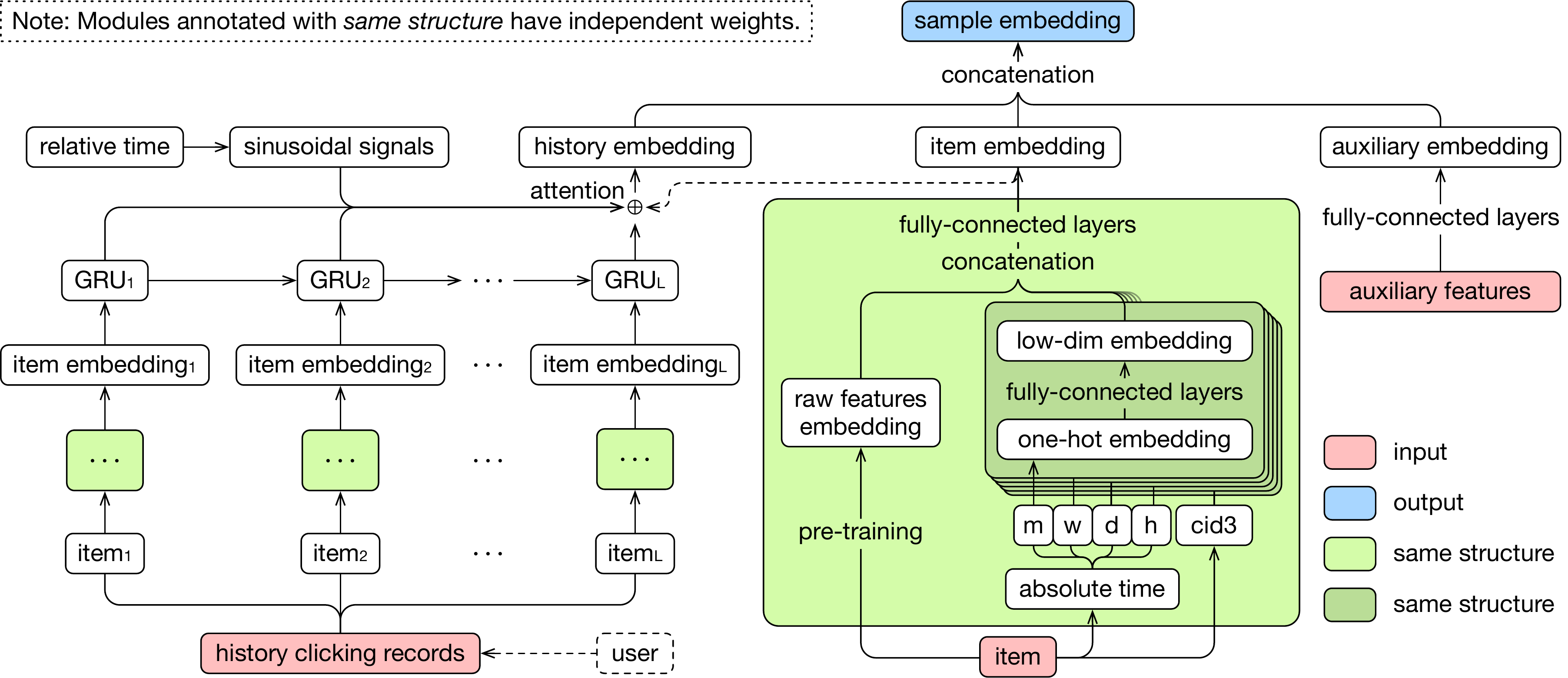}
\caption{Structure of the time-aware attention model}
\label{embedding}
\end{figure*}

The structure of the sample embedding model is illustrated in Figure \ref{embedding}. The sample embedding (blue block in Figure \ref{embedding}) is a concatenation of a history embedding, a (target) item embedding and an auxiliary embedding. 

The structure of the item embedding is a nonlinear mapping to features $(\bm{e}_0^r,t_0,c_0)$ of the target item. The absolute time $t_0$ is spilt into $t_0^m$, $t_0^w$, $t_0^d$ and $t_0^h$ as defined in Section \ref{sec21}. The four time signals, as well as the cid3 $c_0$, are five scalars followed by the same network structure (dark green block in Figure \ref{embedding}) with independent weights. These scalars are converted to one-hot embeddings and then sent to a fully-connected layer for dimensional reduction. The five low dimensional embeddings together with the raw features embedding $\bm{e}_0^r$ are concatenated and converted using fully-connected layers to a $d$-dimensional item embedding, denoted as $\bm{e}_0^i$. The structure of the item embedding is formulated as follows:
\begin{equation}
\begin{split}
&\bm{e}_0^c=\mathcal{R}\big(\bm{W}_c \mathcal{O}(c_0)\big)\,,\\
&\bm{e}_0^t=\mathcal{R}\big(concat\big(\bm{W}_m \mathcal{O}(t_0^m),\bm{W}_w \mathcal{O}(t_0^w),\bm{W}_d \mathcal{O}(t_0^d),\bm{W}_h \mathcal{O}(t_0^h)\big)\big)\,,\\
&\bm{e}_0^i=\mathcal{F}\big(concat\big(\bm{e}_0^r,\bm{e}_0^c,\bm{e}_0^t\big)\big)\,,
\end{split}
\end{equation}
where $\mathcal{O}(\cdot)$ is a projection of the one-hot embedding; $\mathcal{R}(\cdot)$ is a ReLU function; $\mathcal{F}(\cdot)$ is a series of fully-connected layers; $\bm{W}_*$ are trainable matrixes for dimensional reduction; $\bm{e}_0^c$ and $\bm{e}_0^t$ represent the embeddings of the cid3 and the absolute time respectively.

For $L$ historical items in a sample, we get their corresponding item embeddings using the same network structure (light green block in Figure \ref{embedding}) as the embedding structure of the target item. Denote these item embeddings as $\bm{e}_l^i\in\mathbb{R}^{d},l=1\cdots L$. To capture the sequential information, we apply GRU mechanism\cite{DBLP:conf/emnlp/ChoMGBBSB14} to the historical item embeddings, which is formulated as:
\begin{equation}
\begin{split}
&\bm{r}_l=\sigma(\bm{W}_{er}\bm{e}_l^i+\bm{W}_{hr}\bm{h}_{l-1}+\bm{b}_r)\,,\\
&\bm{z}_l=\sigma(\bm{W}_{ez}\bm{e}_l^i+\bm{W}_{hz}\bm{h}_{l-1}+\bm{b}_z)\,,\\
&\widetilde{\bm{h}}_l=tanh\big(\bm{W}_{ec}\bm{e}_l^i+\bm{W}_{hc}(\bm{r}_l\odot\bm{h}_{l-1})+\bm{b}_c\big)\,,\\
&\bm{h}_l=(1-\bm{z}_l)\odot\bm{h}_{l-1}+\bm{z}_l\odot\widetilde{\bm{h}}_l\,,
\end{split}
\end{equation}
where $\sigma(\cdot)$ is a sigmoid function; $\bm{h}_l\in\mathbb{R}^{h}$ is the hidden state; $\bm{r}_l$ is the reset gate controlling the input of the former hidden state $\bm{h}_{l-1}$; $\bm{z}_l$ is the update gate controlling the update ratio; $\bm{W}_{**}$ and $\bm{b}_*$ are trainable variables.

For exploration of each individual historical item, we apply the attention mechanism to the hidden states of GRU, by assigning proper weights to various hidden states. Instead of using the standard attention, given the speciality of CTR prediction, we provide the attention model with $d$-dimensional relative time embeddings. The relative time, specified in seconds, is an attribute of each historical clicking item representing the interval between its clicking time and the exposing time of the target item.
The relative time embedding of the $l$th historical item, denoted as $\bm{e}_l^t$, is calculated as:
\begin{equation}
\begin{split}
&\bm{e}_l^t[2j]=sin\big((t_0-t_l)/10000^{2j/d}\big)\,,\\
&\bm{e}_l^t[2j+1]=cos\big((t_0-t_l)/10000^{2j/d}\big)\,,
\end{split}
\end{equation}
where $2j$ and $2j+1$ are the indexes; $t_0$ and $t_l$  are the absolute time in seconds of the target item and the $l$th historical item respectively; the coefficient 10000 is referring to the positional encodings in \cite{DBLP:conf/nips/VaswaniSPUJGKP17}.

In the attention module, the weighted factor $a_l$ of the $l$th hidden state $\bm{h}_l$ is formulated as:
\begin{equation}
\begin{split}
&u_l=\bm{v}^{\top}tanh(\bm{W}_{h}\bm{h}_l+\bm{W}_{i}\bm{e}_0^i+\bm{W}_{t}\bm{e}_l^t)\,,\\
&a_l=softmax(u_l)\,,
\end{split}
\end{equation}
where $u_l$ is an intermediate variable; $\bm{e}_0^i\in\mathbb{R}^{d}$ is the item embedding of the target item; $\bm{e}_l^t\in\mathbb{R}^{d}$ is the relative time embedding; $\bm{W}_{h}\in\mathbb{R}^{v\times h}$, $\bm{W}_{i}$, $\bm{W}_{e}\in\mathbb{R}^{v\times d}$, $\bm{v}\in\mathbb{R}^{v}$ are trainable variables.

The history embedding $\bm{e}^h$ of the sample, which is the output of the attention module, is a linear weighted sum formulated as:
\begin{equation}
\bm{e}^h=\sum_{l=1}^L a_l\bm{h}_l\,.
\end{equation}

The auxiliary embedding $\bm{e}^a$ is a nonlinear mapping of the auxiliary features described in Section \ref{sec21} using fully-connected layers.

Finally the sample embedding $\bm{e}^s$ (blue block in Figure \ref{embedding}) is a concatenation of the history embedding $\bm{e}^h$, the (target) item embedding $\bm{e}_0^i$ and the auxiliary embedding $\bm{e}^a$, as follows:
\begin{equation}
\bm{e}^s=concat\big(\bm{e}^h,\bm{e}_0^i,\bm{e}^a\big)\,.
\end{equation}

After a user browsing through a recommended item, the user-item sample is labeled as positive or negative based on whether the user clicks the item or not. Given the labels, our model can be learned by minimizing either a pointwise or a pairwise loss. The pointwise loss function is a cross-entropy loss calculated as:
\begin{equation}
\mathcal{L}_{point}=\sum_{s\in\mathcal{T}}\log\sigma\big(f\big(\bm{e}^s(s)\big)\big)+\sum_{s\in\mathcal{T}'}\log\big(1-\sigma\big(f\big(\bm{e}^s(s)\big)\big)\big)\,,
\end{equation}
where $\mathcal{T}$ and $\mathcal{T}'$ are the sets of the whole positive and the negative samples respectively; $f(\cdot)$ is a score function that converts the sample embedding to a scalar score; $\sigma(\cdot)$ is a sigmoid function. 

The pairwise loss function is a marginal hinge loss calculated as:
\begin{equation}\label{pairwise}
\mathcal{L}_{pair}=\sum_{s\in\mathcal{T},s'\in\mathcal{T}'}\big[-f\big(\bm{e}^s(s)\big)+f\big(\bm{e}^{s}(s')\big)+\gamma\big]_+\,,
\end{equation}
where $[\cdot]_+=\max(0,\cdot)$ is the hinge function, and $\gamma$ is the margin.

In CTR prediction tasks, the size of the positives $|\mathcal{T}|$ is far smaller than the size of the negatives $|\mathcal{T}'|$, and thus both loss functions suffer from the imbalanced problem. Because of this, usual works conduct a negative sampling process instead of directly using the whole negative samples. In the next section, to improve the performance of CTR prediction, we propose a GAN-based negative sampling strategy which is adaptable for the both loss functions. We take the pairwise loss for example and provide detailed derivations.

\section{Regularized Adversarial Sampling}
\label{sec3}
In this section we introduce our adversarial negative sampling model which is specifically designed for CTR prediction. The model contains a distance-based discriminator and a probability-based generator. The proposed adversarial sampling model can make use of the guidance brought by the observed negative samples.
\subsection{Sampling Strategy}
\label{sub:sampling_strategy}

The sampling model trains a pairwise loss which is similar to Eq. (\ref{pairwise}). Specifically, at each step, a positive sample $s$ and a negative sample $s'$ are embedded using a time-aware attention module, and are both sent to the discriminator $D$. Networks in the discriminator convert the sample embeddings into scores, denoted as $f_D(\bm{e}_D^s(s))$ and $f_D(\bm{e}_D^s(s'))$ respectively. Then the optimizer in the discriminator updates to increase the gap between the scores. The score $f_D$ measures the attraction of the item to the user. Therefore positive samples tend to have higher scores than the negatives. 

When given a positive sample in each training step, a negative sample needs to be selected for training the pairwise loss function in the discriminator. One straightforward way is to uniformly choose a negative sample from all the negatives, and we call this method \emph{uniform sampling}. Instead of using uniform sampling, we are seeking a better sampling strategy to enhance training efficiency and thus acquire better CTR prediction performance. 

Supposing for a given positive sample $s$, a negative sample $s'$ is sampled following a distribution conditioned on $s$. In our adversarial sampling model, We use a generator $G$ to fit this conditional distribution, denoted as $p_G(s'|s)$. In the generator, we adopt another time-aware attention model for each sample to get the sample embedding, denoted as $\bm{e}_G^s$. The attention model in the generator shares the same structure with the attention model in the discriminator, but has independent network weights. The conditional distribution $p_G(s'|s)$ is expressed by an explicit union function of $\bm{e}_G^s(s)$ and $\bm{e}_G^s(s')$, which will be analyzed in later part of this section. 

Referring to Eq. (\ref{pairwise}), the objective of the discriminator can be formulated as minimizing the following hinge loss function:
\begin{equation}
\mathcal{L}_D = \sum_{s\in\mathcal{T}}\big[-f_D\big(\bm{e}_D^s(s)\big)+f_D\big(\bm{e}_D^s(s')\big)+\gamma\big]_+,\;s'\sim p_G(s'|s)\,,
\end{equation}
where $\mathcal{T}$ is the set of the positive samples. Minimizing $\mathcal{L}_D$ tends to increase $f_D(\bm{e}_D^s(s))$ and decrease $f_D(\bm{e}_D^s(s'))$ to ensure a gap $\gamma$.

In order to design a better sampling strategy $p_G(s'|s)$, we take two aspects into consideration. Firstly, at a training step, if the selected negative sample $s'$ has a low score, the gap between $f_D(\bm{e}_D^s(s))$ and $f_D(\bm{e}_D^s(s'))$ may close to or already larger than $\gamma$ before any update, which leads to a useless training step. This will cause a low data efficiency and therefore affect the performance. Secondly, CTR prediction is a largely imbalanced classification problem. One-sided selection\cite{DBLP:conf/micai/BatistaCM00}, a typical neighbor-based imbalanced classification approach, focuses on the boundary negatives and removes redundant or interferential majorities, achieving a better classification performance. Inspired by this, we consider a negative sample in CTR data to be redundant regarding to a given positive sample, if the negative sample locates too far away from the positive sample in some certain embedding spaces. The optimizer will acquire an unsatisfactory performance if it pays too much attention on the redundant negative samples, even though they have high scores.

To sum up, when given a positive sample $s$, a negative sample $s'$ provided by the generator should have two desirable properties:
\leftmargini=7mm
\begin{itemize}
\setlength\itemsep{0em}
\item Competitive: $s'$ should have a high score, so as to be a strong impetus when training the discriminator.
\item Correlative: Enough similarity needs to be ensured between the embeddings of $s$ and $s'$. We experimentally find that using Euclidean distances on the embeddings calculated in the \emph{discriminator embedding space} works better. 
\end{itemize}

Detailed experimental results in subsection \ref{sec:indepth} compare the individual performance of either one of the properties and the overall performance with a good cooperation of the two properties.

\begin{figure}[ht]
\centering
\includegraphics[scale=0.26]{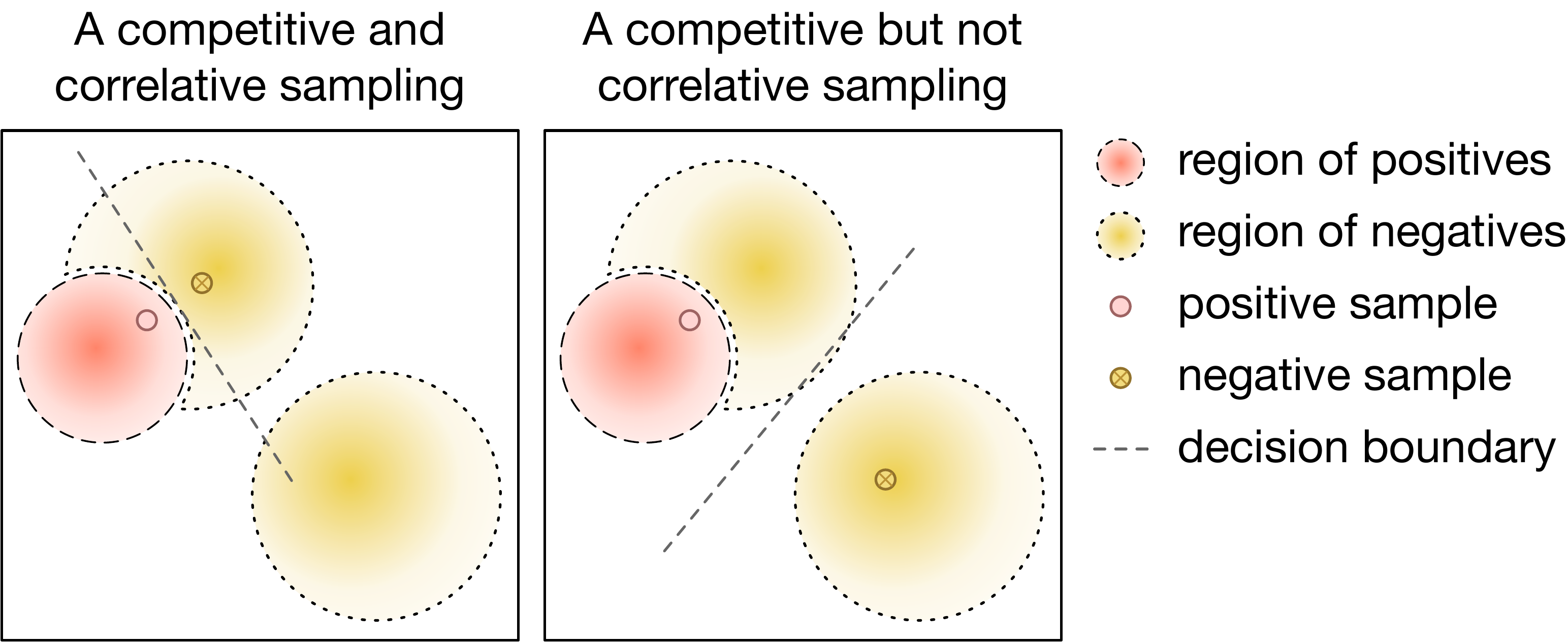}
\caption{Effect visualization of the correlative sampling}
\label{correlative}
\end{figure}

We visualize the influence to be correlative in Figure \ref{correlative}, where the color shade reflects the score of the sample. A sample with darker shade in either positive region or negative region is more tend to be positive in the view of the discriminator. In each sub-figure, a negative sample with a high score is selected regarding to a given positive sample. The difference is, the first sampling seeks a tighter and more effective decision boundary compared with the second, giving the credit to the correlative negative sampling.

Denote the item and history embedding functions in the discriminator as $\bm{e}_D^i$ and $\bm{e}_D^h$ respectively. In order to select a negative sample $s'$ which keeps correlative to the positive sample $s$. We use Euclidean distances in the discriminator embedding space as a penalty $p(s,s')$, formulated as:
\begin{align}\label{penalty}
p(s,s')=\lambda_i\big\|\bm{e}_D^i(s)-\bm{e}_D^i(s')\big\|_2+\lambda_h\big\|\bm{e}_D^h(s)-\bm{e}_D^h(s')\big\|_2\,,
\end{align}
where $\lambda_i,\lambda_h>0$ are penalty coefficients for the item embedding penalty and the history embedding penalty respectively. 

The objective of the generator is designed to maximize the expectation of scores of the selected negative samples with penalties for correlative restriction. The loss function can be written as:
\begin{align}
\mathcal{L}_G=\sum_{s\in\mathcal{T}}\underbrace{\mathbb{E}_{s'\sim p_G(s'|s)}\big[f_D\big(\bm{e}_D^s(s')\big)-p(s,s')\big]}_{\text{denote as }\mathcal{J}_G(s)}\,.
\end{align}

A typical technique to calculate the gradient of the function with expectations is to adopt the idea of policy gradient based reinforcement learning (REINFORCE)\cite{DBLP:journals/ml/Williams92, DBLP:conf/aaai/YuZWY17}. Denote the parameters in the generator as $\theta_G$. The gradient of $\mathcal{J}_G(s)$ is derived as: 

\begin{align}\label{pg}
\hspace{-0.07in}
\nabla_{\theta_G} \mathcal{J}_G(s)&=\nabla_{\theta_G}\mathbb{E}_{s'\sim p_G(s'|s)}\big[f_D\big(\bm{e}_D^s(s')\big)-p(s,s')\big]\nonumber\\
&=\sum_{n=1}^N\nabla_{\theta_G} p_G(s'_n|s)\big[f_D\big(\bm{e}_D^s(s'_n)\big)-p(s,s'_n)\big]\nonumber\\
&=\sum_{n=1}^N p_G(s'_n|s)\nabla_{\theta_G} \log p_G(s'_n|s)\big[f_D\big(\bm{e}_D^s(s'_n)\big)-p(s,s'_n)\big]\nonumber\\
&=\mathbb{E}_{s'\sim p_G(s'|s)}\big\{\nabla_{\theta_G}\log p_G(s'|s)\big[f_D\big(\bm{e}_D^s(s')\big)-p(s,s')\big]\big\}\nonumber\\
&\simeq \frac{1}{K}\sum_{k=1}^K\nabla_{\theta_G}\log p_G(s'_k|s)\big[f_D\big(\bm{e}_D^s(s'_k)\big)-p(s,s'_k)\big]\,.
\end{align}

The approximation in Eq. (\ref{pg}) is based on Monte Carlo method. To update the generator, at each step, we sample a mini-batch of positives and $K$ corresponding negative samples for each positive sample. With the REINFORCE terminology, the term $[f_D(\bm{e}_D^s(s'_k))-p(s,s'_k)]$, denoted as $r(s'_k)$, acts as the reward for the policy $p_G(s'_k|s)$ which takes an action $s'_k$ given the environment state $s$. 

Based on the above description, we can utilize the strong guidance brought by the observed negatives, by adversarially selecting the competitive negatives with embedding distance regularizations. 

Our GAN-based sampling strategy is specifically designed for CTR prediction tasks or some of the recommender tasks where observed negative samples are available. We further unify our sampling strategy with the previous strategy illustrated in Figure \ref{structure_comparison}, where no observed negatives are available and the nonpositive samples (randomly user-item combinations which are not positive) are used to train against the positive samples. In this case, we fix the user part of each positive sample, and combine the user to an item that are not interacted with the user to form a negative sample. With the same user part, the history embedding penalty in Eq. (\ref{penalty}) is zero, thus $r(s'_k)$ can degenerate to an easier form. Summarizing the both conditions, the reward $r(s'_k)$ is formulated as:
\begin{equation}\label{reward}
\begin{split}
r(s'_k)=\left\{
\begin{array}{ll}
f_D\big(\bm{e}_D^s(s'_k)\big)-p(s,s'_k)&\text{with negatives}\\[2mm]
f_D\big(\bm{e}_D^s(s'_k)\big)-\lambda_i\big\|\bm{e}_D^i(s)-\bm{e}_D^i(s'_k)\big\|_2&\text{without negatives}\\
\end{array}\right.
\end{split}\hspace{-0.07in}.
\end{equation}
As a common optimization in policy gradient to reduce variance, we can subtract a baseline $b$ from the reward, where $b$ equals to the average reward of the mini-batch $\mathcal{T}_{batch}$. The baseline $b$ updates after each training step by the following equation:
\begin{equation}
\label{baseline}
b=\frac{1}{|\mathcal{T}_{batch}|}\sum_{s\in\mathcal{T}_{batch}}\frac{1}{K}\sum_{k=1}^K r(s'_k)\,.
\end{equation}

With the technique of stochastic optimization, in a mini-batch $\mathcal{T}_{batch}$, the gradient of loss function $\mathcal{L}_G$ in the generator is:
\begin{equation}\label{lg}
\nabla_{\theta_G} \mathcal{L}_G\simeq
\sum_{s\in\mathcal{T}_{batch}}\frac{1}{K}\sum_{k=1}^K\log p_G(s'_k|s)\big[r(s'_k)-b\big]\,,
\end{equation}
where the baseline $b$ is obtained from the previous mini-batch. 

The sampling policy $p_G(s'|s)$ for the negative sample $s'$ regarding to a positive sample $s$, is modeled as a union function of the generator sample embeddings $\bm{e}_G^s(s)$ and $\bm{e}_G^s(s')$:
\begin{equation}\label{softmax}
p_G(s'|s)=softmax_{s'\in Neg(s)}\frac{\bm{e}_G^s(s')^{\top}\bm{e}_G^s(s)}{T\big\|\bm{e}_G^s(s')\big\|_2}\,,
\end{equation}
where $Neg(s)$ is the set of candidate negative samples for sample $s$; $T$ is a temperature for sensitivity control; dividing $\|\bm{e}_G^s(s')\|_2$ is to eliminate the unfairness brought by the scale of $\bm{e}_G^s(s')$. 

As the negative samples in practical CTR tasks are noisy, negative samples with the highest scores are likely to be false negatives, i.e. neglected positive samples, which will impact the training performance. To tackle this issue, we generate the set $Neg(s)$ by uniformly sampling $C$ negative samples as candidates, instead of the whole negatives, where $C$ is a hyperparameter. Denote the size of the positives is $|\mathcal{T}|$. For each positive sample $s$, calculating its corresponding policy needs $O(C)$ time complexity, and the total policy calculation expense in an epoch is $O(|\mathcal{T}|\cdot C)$. Such a complexity can be reduced to $O(|\mathcal{T}|\cdot\log C)$ with hierarchical softmax\cite{DBLP:conf/nips/MikolovSCCD13}. 

\begin{algorithm}[htb]   
\caption{Training the adversarial sampling network}   
\label{full_al}   
\begin{algorithmic}[1] 
\REQUIRE
positive samples $\mathcal{T}$ and negative samples $\mathcal{T}'$
\ENSURE adversarially trained discriminator
\STATE Pre-train the discriminator $D$ and the generator $G$
\REPEAT 
\STATE Sample a mini-batch positive samples $\mathcal{T}_{batch}\in\mathcal{T}$;
\FOR{each positive sample $s\in\mathcal{T}_{batch}$}
\STATE Uniformly select $C$ candidate negative samples from $\mathcal{T}'$
\STATE In $G$, sample a negative sample $s'$ from the $C$ candidates according to the policy $p_G(s'|s)$ in Eq. (\ref{softmax})
\STATE In $D$, optimize $\mathcal{L}_D$ by a stochastic gradient descent
\STATE In $D$, calculate the score $f_D(s')$ and the reward $r(s')$ according to Eq. (\ref{reward})
\STATE In $G$, optimize $\mathcal{L}_G$ by a stochastic gradient ascent according to Eq. (\ref{lg})
\ENDFOR
\UNTIL convergence 
\end{algorithmic}  
\end{algorithm}

GAN can suffer from the mode collapse issue where the generator collapses to some certain modes\cite{DBLP:journals/corr/ArjovskyCB17}. In our sampling process, the generator relies on $p_G(s'|s)$ to sample the negatives, and the collapse of $p_G(s'|s)$ to some certain negatives will overfit on these negatives. Thus we should balance the exploration process referring to traveling through the negatives, and the exploitation process referring to sampling the negatives unfairly regarding to the rewards. 

In Eq. (\ref{softmax}), the sensitivity of softmax can be adjusted by adjusting the temperature $T$. A larger $T$ will lead to a lower sampling sensitivity, which encourages the generator to explore, and a lower $T$ encourages the generator to exploit. The generator should explore at the early training
to meet more negative samples, while exploit later to take
advantage of the adversarial sampling. Thus we give a large initial
temperature and decay it epoch by epoch with a decay rate. Sensitivity analysis of $T$ can be seen in Section \ref{sec:indepth}.

We experimentally set the parameter $K$ in Eq. (\ref{pg}) to one and find it sufficient to deliver promising results. The main framework of our adversarial training algorithm is described in Algorithm \ref{full_al}.

\begin{table*}[h]\centering
\caption{Summary of the data structure (Year: 2018)}
\label{data}
\resizebox{\linewidth}{!}{
\begin{tabular}{c|c|c|c|c|c|c|c|c|c|c}
\toprule
\multirow{2}*{Dataset}  &\multirow{2}*{Items} &\multirow{2}*{Categories} &\multicolumn{4}{c|}{Training Data} &\multicolumn{4}{c}{Testing Data} \\ 
&&&Date &Positives &Negatives &CTR &Date &Positives & Negatives &CTR\\
\midrule
\multirow{1}*{In-station-Sep.} &1,680,735 &3,137 &Sep.14th &126,016 &8,233,656 &1.507\% &Sep.15th &138,676 &9,173,364 &1.489\%\\
\multirow{1}*{Out-station-Jul.}&3,411,957 &4,574 &Jul.5th$\sim$Jul.6th &35,262 &2,868,936 &1.214\% &Jul.7th &17,630 &1,352,200 &1.287\%\\
\multirow{1}*{Out-station-May.} &3,734,557 &4,389 &May.6th &18,309 &1,502,377 &1.204\% &May.7th &18,780 &1,514,364 &1.220\%\\
\bottomrule
\end{tabular}}
\end{table*}

\subsection{Output Calibration}
\label{subsec33}
The proposed adversarial sampling strategy brings better relative CTR values which benefit the ranking process in CPC advertising. But during the negative sampling, the proportion of the positives and the negatives in the constructed training data will not match the real data proportion. Such mismatching will lead to an inaccurate absolute CTR estimates which is bad for the bidding process. Inspired by the calibration method in \cite{DBLP:conf/kdd/LeeODL12}, we adopt a similar output calibration by taking into account the real data proportion.

The real CTR of a sample $s$ for training can be expressed as:
\begin{equation}
\text{CTR}(s)=p\big(Y(s)=1|f_D\big(\bm{e}_D^s(s)\big)\big)\,,
\end{equation}
where $Y(s)=1$ indicates that the sample $s$ is a positive sample.

We apply a sigmoid function on the score $f_D(\bm{e}_D^s(s))$ for $[0,1]$ normalization, and denote the normalized score as $\sigma(s)$. We divide the region $[0,1]$ into $n$ equal-sized buckets, where $0\le v_1<v_2<,\cdots,<v_{n+1}\le1$ and $n$ is large enough. Assuming that $\sigma(s)$ locates in $[v_j,v_{j+1}),~j=1,\cdots,n$, an approximation of $\text{CTR}(s)$ is:
\begin{align}\label{ctr}
\text{CTR}(s)\simeq p\big(Y(s)=1|v_j\le \sigma(s)<v_{j+1}\big)\,,
\end{align}

Denote the right part of Eq. (\ref{ctr}) as $p(v_j),~j=1,\cdots,n$. An approximation of $p(v_j)$ is the proportion of the positive training samples in all training samples, written as:
\begin{align}\label{pvj}
p(v_j)\simeq\frac{\#\text{ Positive training samples with } \sigma(s)\in[v_j,v_{j+1})}{\#\text{ All training samples with } \sigma(s)\in[v_j,v_{j+1})}\,.
\end{align}

An isotonic regression algorithm should be applied to keep $p(v_j)$ monotonically increasing with $j=1,\cdots,n$. We experimentally choose Pool Adjacent Violators Algorithm\cite{Leeuw2009Isotone} for regression.

After obtaining the estimates of $p(v_j),~j=1,\cdots,n$. For a single sample $\hat{s}$ in the testing data, its CTR is calibrated by:
\begin{align}
\text{CTR}(\hat{s}) &\simeq p\big(Y(\hat{s})=1|v_j\le \sigma(\hat{s})<v_{j+1}\big)\nonumber\\
&=\alpha p(v_j)+(1-\alpha)p(v_{j+1})\,,
\end{align}
where we suppose $\sigma(\hat{s})\in[v_j,v_{j+1})$, and $\alpha=\frac{v_{j+1}-\sigma(\hat{s})}{v_{j+1}-v_j}$.

The final calibrated CTR value of the testing data is the average calibrated CTR of all testing samples. Experimental results of the output calibration are shown in Section \ref{sub:results}.

\section{Experiments}
\label{sec4}
We conduct experiments on three real-world industrial datasets provided by JD.COM, one of which is from in-station advertising places (ads are displayed in the JD mobile app) and the other two are from out-station advertising places (ads are displayed in the third-party platforms). The datasets we choose have various magnitudes and CTR values for distinction considerations. A summary of the datasets is provided in Table \ref{data}.

We use AUC (area under the receiver operating characteristic curve) for performance evaluation. AUC is a commonly-used evaluation metric for CTR prediction, which measures the quality of the order by ranking all the samples with predicted CTR\cite{DBLP:journals/prl/Fawcett06}.

Referring to \cite{DBLP:conf/icml/YanLXH14, DBLP:conf/kdd/ZhouZSFZMYJLG18} , we adopt RelaImpr metric to measure the relative enhancement of AUC scores over base models. As the AUC score of a random guess is 0.5, RelaImpr is defined as:
\begin{equation}
\text{RelaImpr}=\left(\frac{\text{AUC of measured model}-0.5}{\text{AUC of base model}-0.5}-1\right)\times100\%\,.
\end{equation}

\subsection{Baselines}
For verifying the time-aware attention model, we select the following base models for comparison, and we uniformly adopt the regularized adversarial sampling for the sampling process.
\leftmargini=7mm
\begin{itemize}
\setlength\itemsep{0em}
\item 
{\bf Two-layer GRU}\cite{DBLP:conf/recsys/DonkersL017}: We implement a two-layer GRU network to represent users' historical records.
\item 
{\bf DIN}\cite{DBLP:conf/kdd/ZhouZSFZMYJLG18}: DIN adopts an attention-based model (without GRU) for activating related user behaviors and obtains the relative interests to a target item.
\item 
{\bf GRU Attention}: We discard the temporal components of our time-aware attention model, using a two-layer GRU to model users’ sequential behaviors and a regular attention module for activating inner relations between items. 

\end{itemize}

For verifying the regularized adversarial sampling (rGAN), we select the following models for comparison, and we uniformly adopt the time-aware attention model for the embedding process.
\leftmargini=7mm
\begin{itemize}
\setlength\itemsep{0em}
\item
{\bf Logistic Regression}\cite{DBLP:conf/kdd/McMahanHSYEGNPDGCLWHBK13}: Logistic regression is an ordinary pointwise training method which uses all available samples. 
\item
{\bf 1:5 Under Sampling}\cite{DBLP:conf/icdm/LiuWZ06,DBLP:conf/kdd/HePJXLXSAHBC14}: Under sampling is a common sampling method to balance distributions for CTR prediction. The ratio 1:5 is a tuned choice that realizes better results among various under sampling ratio for our datasets. 
 \item 
{\bf User-fixed Sampling}\cite{DBLP:journals/corr/abs-1708-03993}: User-fixed sampling is a regular sampling method for CTR prediction which selects negative samples that share the same users with the positives. Uniform sampling is used as a complementary choice if there are not enough negative samples that meet requirements.
\item 
{\bf Uniform Sampling}\cite{DBLP:conf/kdd/HePJXLXSAHBC14}: Uniform sampling, where the negatives are selected uniformly given any positive sample, is a commonly-used CTR sampling method, and can be seen as a degradation of our rGAN strategy when $T$ is large or $C=1$.
\item
{\bf IRGAN}\cite{DBLP:conf/sigir/WangYZGXWZZ17}: IRGAN is a state-of-the-art model which applies adversarial sampling to select non-interactive user-item samples to train against the positive samples. However, IRGAN cannot use the information of the practically observed negative samples in CTR prediction tasks.
\item
{\bf IRGAN++}: We modify IRGAN to sample from the practically observed negatives. As directly sampling from the negatives with IRGAN fails to consider the correlation (illustrated in Figure \ref{correlative}) between samples, we explore some modifications of the original IRGAN by narrowing down the sampling scope. We denote the modification with the best result as IRGAN++: given a positive sample, IRGAN is applied to sample from the negatives that share the same target item with the positive sample; and if there are not sufficient negative samples, we sample from the negatives of which the target items belong to the same category as the target item of the positive sample.

\end{itemize}

\subsection{Implementation Details}
\label{details}
The structural details of the embedding networks illustrated in Figure \ref{embedding} are described as follows: 
\leftmargini=7mm
\begin{itemize}
\setlength\itemsep{0em}
\item\label{item_embedding}
Item embedding: For an item, we convert its cid3 and four absolute time signals to five one-hot embeddings, each of which is further sent to a fully-connected layer with tanh activation function. Every obtained output has eight dimensions, and is concatenated with the item's 50-dimensional raw features embedding obtained by pre-training using Telepath\cite{DBLP:conf/aaai/WangXWLHHY18}. The fully-connected network between the concatenation and the item embedding has three 90-dimensional layers, with ReLU for the hidden layers and tanh for the output layer.
\item 
History embedding: Each historical record in a sample contains the user's 10 latest clicking items, and their corresponding item embeddings are obtained using the same structure described in (\ref{item_embedding}). The number 10 is selected by experimental results. These item embeddings are sent to a two-layer GRU network with 90 RNN size as inputs. The dimension of the relative time embedding is 90. We further apply the time-aware attention module on GRU. The history embedding, which is the output of the attention, has 90 dimensions. 

\item
Auxiliary embedding: We apply a one-layer fully-connected network with an eight-dimensional output layer on the auxiliary features. The layer uses tanh as the activation function.

\end{itemize}

In the discriminator, the score net is a linear single-layer fully-connected network with one-dimensional outputs.

We use Adam\cite{DBLP:journals/corr/KingmaB14} for both trainings in discriminator and generator. Parameters follow $\alpha=0.001$, $\beta_1=0.9$, $\beta_2=0.999$, $\epsilon=10^{-8}$.

The model is trained for 50 epochs, where the first 25 are shown in results. For each sample batch, we set one step for training either the discriminator or the generator. We adjust the batch size to keep 30 steps per epoch. The initial learning rate for the discriminator and the generator are respectively set to $0.02$ and $0.01$, which both decay $50\%$ every 10 epochs. $K$ in Eq. (\ref{lg}) is set to one. The margin of the loss function is $1.0$. The penalty coefficients for the item embedding and the history embedding are $3.0$ and $5.0$ respectively. The candidate size $C$ is set to $35$ for the In-station-Sep. dataset and set to $20$ for the other two datasets. The initial temperature $T$ is constantly set to $20.0$, with a decay rate 0.98.

\subsection{Comparison Results}
\label{sub:results}
We tune key hyperparameters individually for each baseline method, and collect the results in two aspects: comparison of the time-aware attention with other embedding models, and comparison of the regularized adversarial sampling with other sampling strategies.

\begin{figure}[h]
  \centering
\hspace{-0.185in}
  \subfigure[In-station-Sep. dataset]{\includegraphics[height=1.3in,width=1.68in]{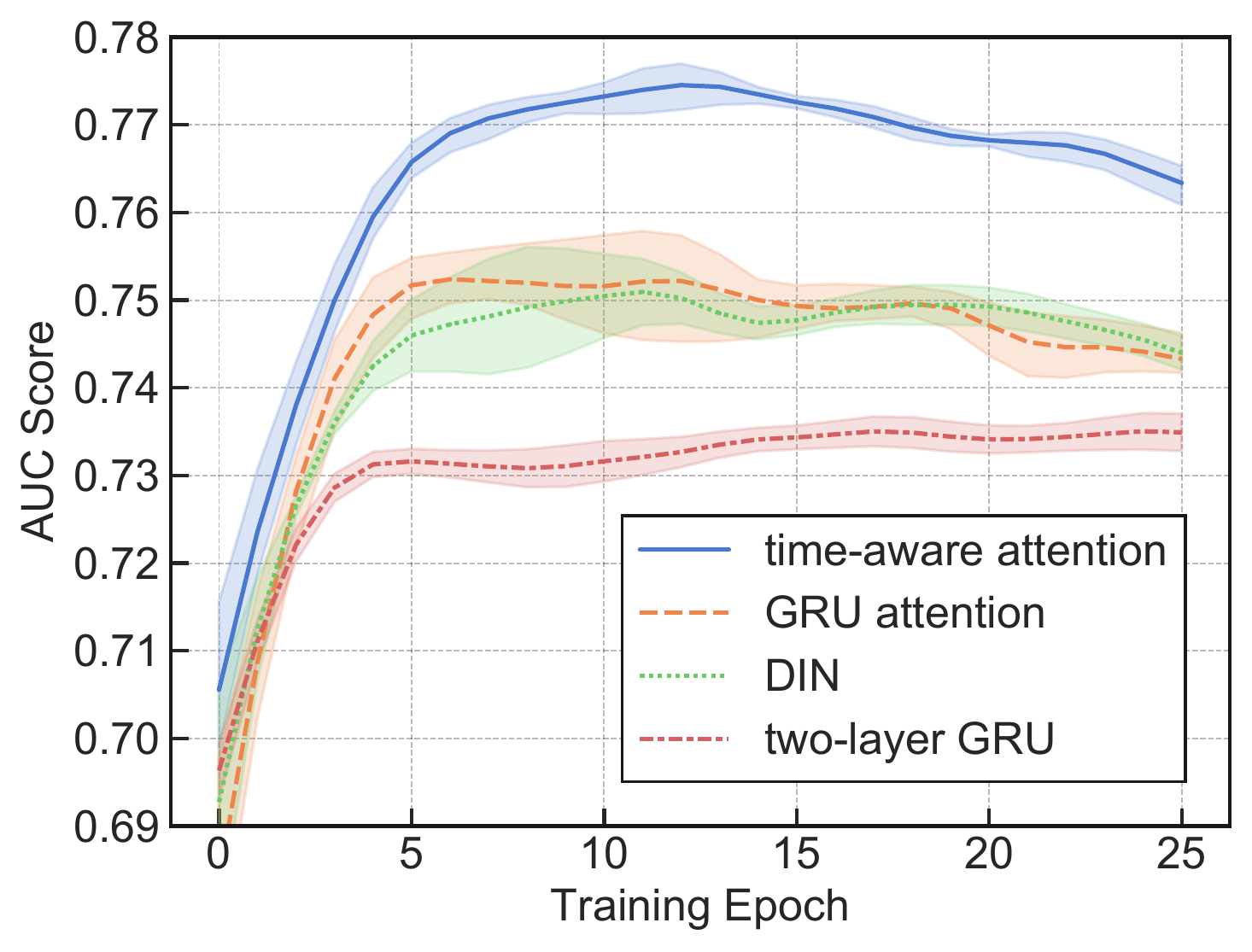}}
\hspace{-0.15in}
  \subfigure[Out-station-Jul. dataset]{\includegraphics[height=1.3in,width=1.68in]{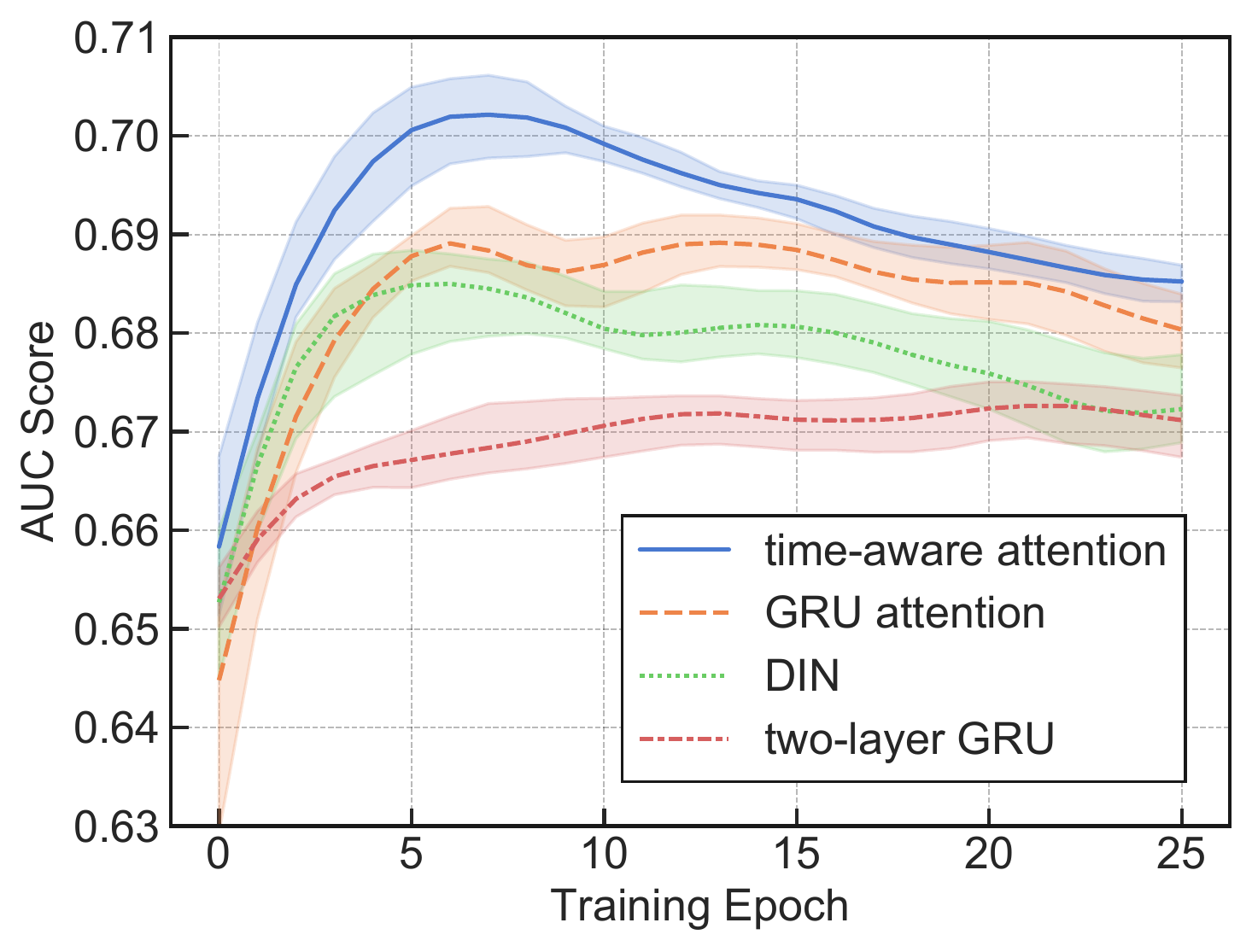}}
\hspace{-0.1in}
  \caption{Performance of various embedding models}
  \label{performance_att}
\end{figure}

Figure \ref{performance_att} illustrates the AUC scores of the time-aware attention and several embedding baselines for two datasets. For each model, we plot the corresponding average curves of five results. We adopt the same regularized adversarial sampling for all the models so as to observe the effect brought by the time-aware attention individually.

Table \ref{table_att} shows the average optimal AUC scores of various embedding models for the three datasets. Observe that the proposed embedding model itself brings $8\%$$\sim$$9\%$ relative AUC improvements. The comparison between the time-aware attention and GRU attention proves the effect of the temporal components.

\begin{table}[h]\centering
\caption{Results of various embedding models}
\label{table_att}
\resizebox{\linewidth}{!}{
\begin{tabular}{c|cc|cc|cc}
\toprule
\multirow{2}*{Model} &\multicolumn{2}{c|}{in-station-Sep.}&\multicolumn{2}{c|}{out-station-Jul.}&\multicolumn{2}{c}{out-station-May.}\\
&AUC&RelaImpr&AUC&RelaImpr&AUC&RelaImpr\\
\midrule
Two-layer GRU&0.7350&-6.33\%&0.6726&-6.70\%&0.6271&-4.36\%\\
DIN*&0.7509&0.00\%&0.6850&0.00\%&0.6329&0.00\%\\
GRU Attention&0.7523&0.56\%&0.6892&2.27\%&0.6322&-0.53\%\\
\textbf{Time-aware Attention}&\textbf{0.7745}&\textbf{9.41\%}&\textbf{0.7021}&\textbf{9.24\%}&\textbf{0.6439}&\textbf{8.28\%}\\
\bottomrule
\end{tabular}}
      \begin{tablenotes}
        \footnotesize
        \item[1] In each table, * indicates the baseline model for calculating RelaImpr.
        \item[1] rGAN sampling is applied for all models in Table \ref{table_att}.
      \end{tablenotes}
\end{table}

\begin{figure}[h]
  \centering
\hspace{-0.185in}
  \subfigure[In-station-Sep. dataset]{\includegraphics[height=1.3in,width=1.68in]{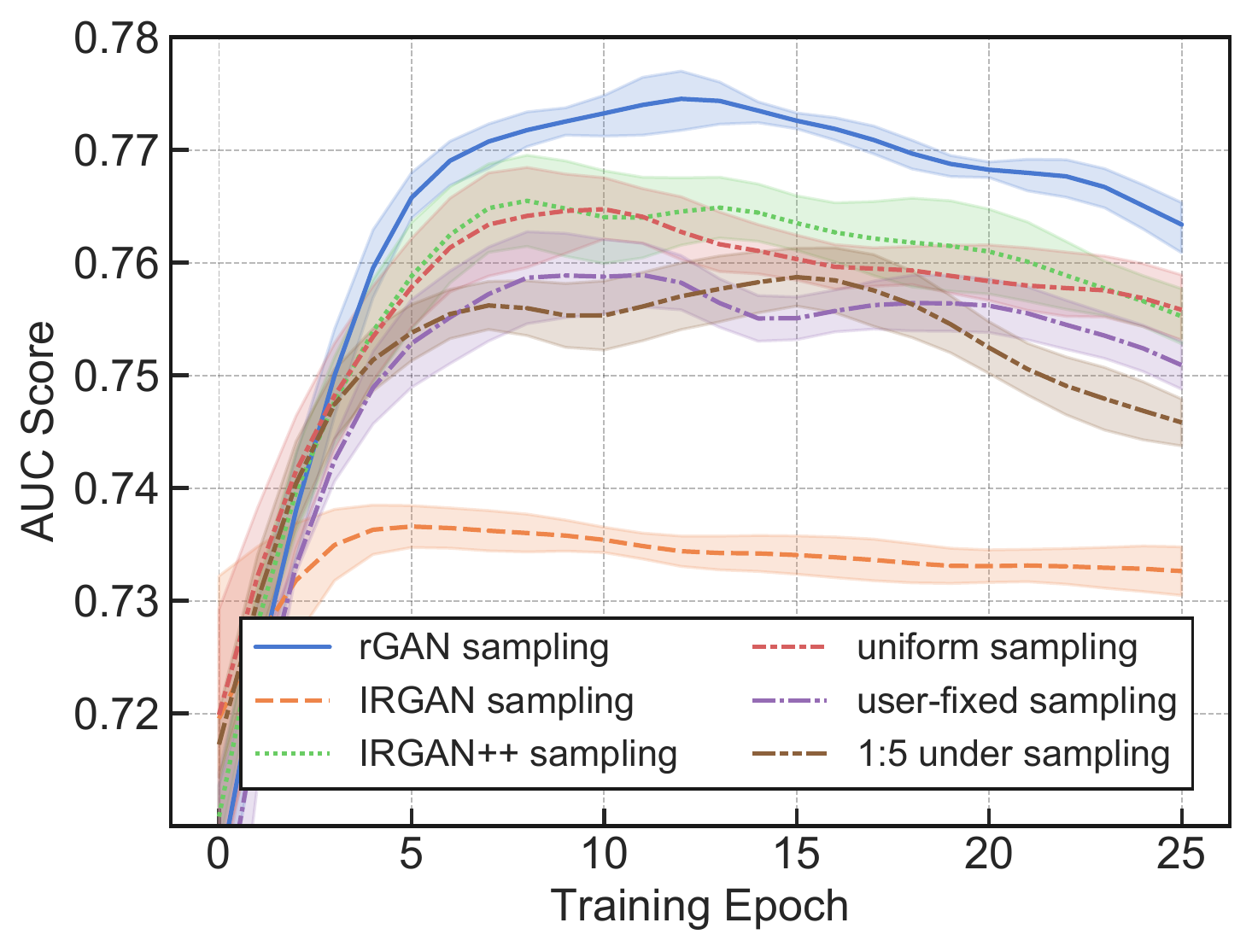}}
\hspace{-0.15in}
  \subfigure[Out-station-Jul. dataset]{\includegraphics[height=1.3in,width=1.68in]{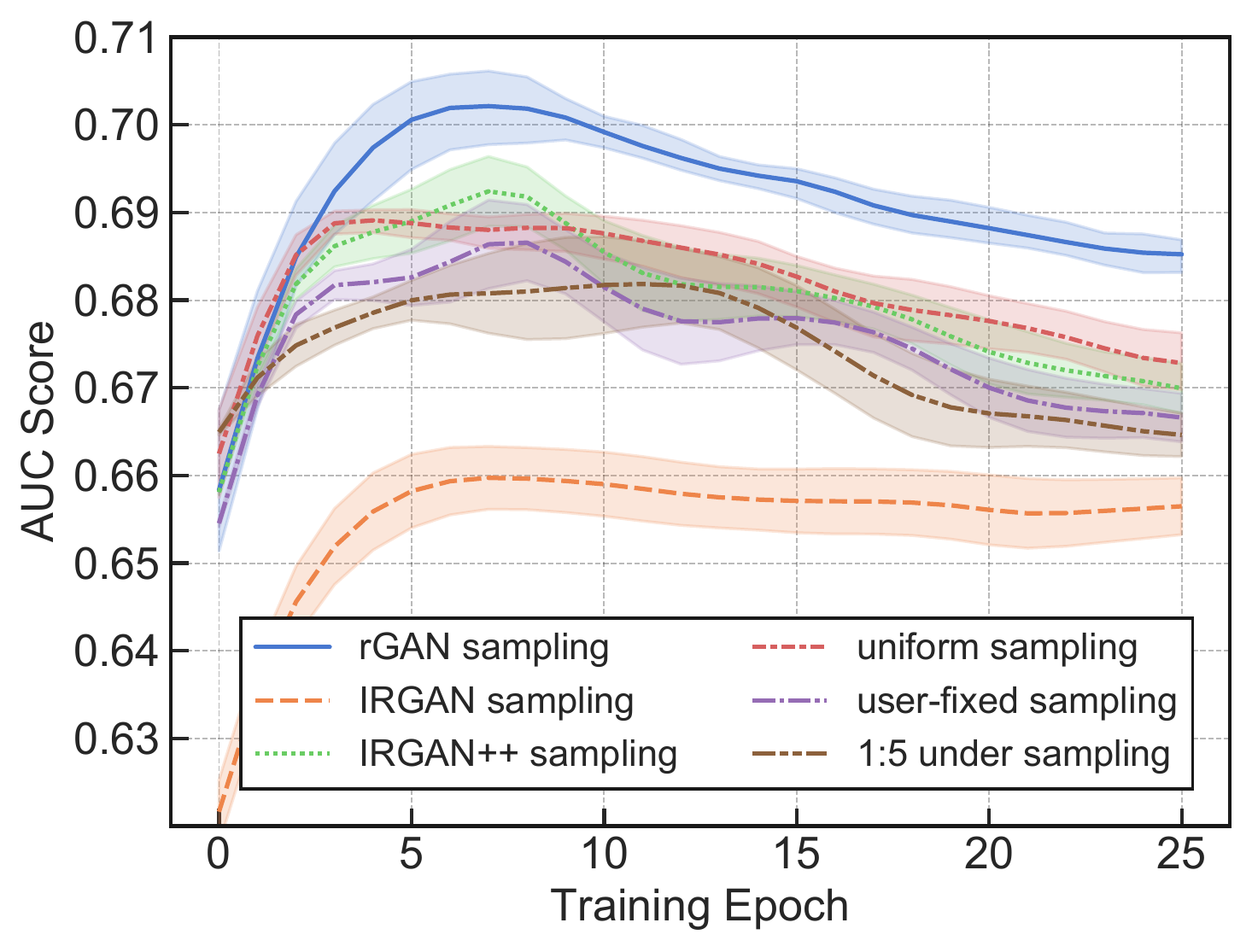}}
\hspace{-0.1in}
  \caption{Performance of various sampling models}
  \label{performance_gan}
\end{figure}

We conduct experiments to verify the effect of regularized adversarial (rGAN) sampling. For each model, we adopt the same time-aware attention model for the embedding process, aiming to observe the individual effect of rGAN sampling. Figure \ref{performance_gan} shows six pairwise training methods, where rGAN sampling model promotes the training performance with better AUC scores.

Table \ref{table_sam} shows the average optimal AUC scores of one pointwise model and six pairwise sampling models on three datasets. During the training process, we find that with proper parameter settings, rGAN stably outperforms the other models. It is worth mention that the rGAN sampling is not only suitable for our time-aware attention model, but also can help  to improve the AUC scores for other embedding models, such as the GRU model and DIN.

\begin{table}[h]\centering
\caption{Results of various sampling models}

\label{table_sam}
\resizebox{\linewidth}{!}{
\begin{tabular}{c|cc|cc|cc}
\toprule
\multirow{2}*{Model} &\multicolumn{2}{c|}{in-station-Sep.}&\multicolumn{2}{c|}{out-station-Jul.}&\multicolumn{2}{c}{out-station-May.}\\
&AUC&RelaImpr&AUC&RelaImpr&AUC&RelaImpr\\
\midrule
Logistic Regression&0.7643&0.15\%&0.6790&-5.34\%&0.6251&-7.81\%\\
\midrule
1:5 Under Sampling&0.7587&-1.97\%&0.6818&-3.86\%&0.6270&-6.41\%\\
User-fixed Sampling&0.7589&-1.89\%&0.6866&-1.32\%&0.6379&1.62\%\\
Uniform Sampling*&0.7639&0.00\%&0.6891&0.00\%&0.6357&0.00\%\\
\midrule
IRGAN Sampling&0.7366&-10.34\%&0.6597&-15.55\%&0.6165&-14.15\%\\
IRGAN++ Sampling&0.7655&0.61\%&0.6924&1.75\%&0.6380&1.69\%\\
\textbf{rGAN Sampling}&\textbf{0.7745}&\textbf{4.02\%}&\textbf{0.7021}&\textbf{6.87\%}&\textbf{0.6439}&\textbf{6.04\%}\\
\bottomrule
\end{tabular}}
\begin{tablenotes}
        \footnotesize
        \item[1] Time-aware attention is applied for all models in Table \ref{table_sam}.
      \end{tablenotes}
\end{table}

As either of the previous results shows the individual effect of the time-aware attention or the regularized adversarial sampling, we now provide their overall improvements in Table \ref{table_overall}. We make comparison between our overall model and DIN with uniform sampling. As shown in the table, our model acquires $11\%$$\sim$$15\%$ relative improvements of the AUC scores, which is 
more remarkable than many state-of-the-art CTR prediction works like \cite{DBLP:conf/ijcai/GuoTYLH17, DBLP:conf/kdd/ZhouZSFZMYJLG18} in real-world datasets. 

\begin{table}[h]\centering
\caption{Overall results of the proposed work}
\label{table_overall}
\resizebox{\linewidth}{!}{
\begin{tabular}{c|cc|cc|cc}
\toprule
\multirow{2}*{Model} &\multicolumn{2}{c|}{in-station-Sep.}&\multicolumn{2}{c|}{out-station-Jul.}&\multicolumn{2}{c}{out-station-May.}\\
&AUC&RelaImpr&AUC&RelaImpr&AUC&RelaImpr\\
\midrule
DIN+Uniform*&0.7405&0.00\%&0.6754&0.00\%&0.6296&0.00\%\\
\textbf{Ours}&\textbf{0.7745}&\textbf{14.14\%}&\textbf{0.7021}&\textbf{15.22\%}&\textbf{0.6439}&\textbf{11.03\%}\\
\bottomrule
\end{tabular}}
\end{table}

\begin{figure}[h]
  \centering
\hspace{-0.185in}
  \subfigure[$p(v_j)$ and its isotonic fitting]{\includegraphics[height=1.05in,width=1.68in]{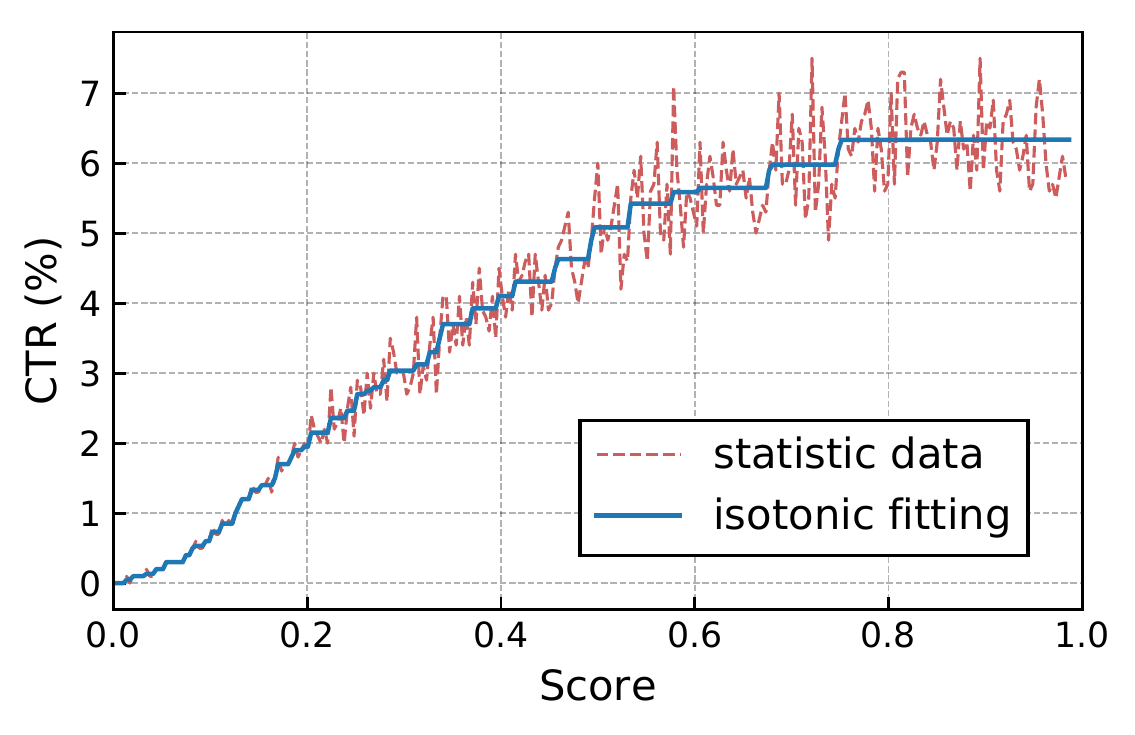}}
\hspace{-0.15in}
  \subfigure[Calibrated results w.r.t. bucket numbers]{\includegraphics[height=1.05in,width=1.68in]{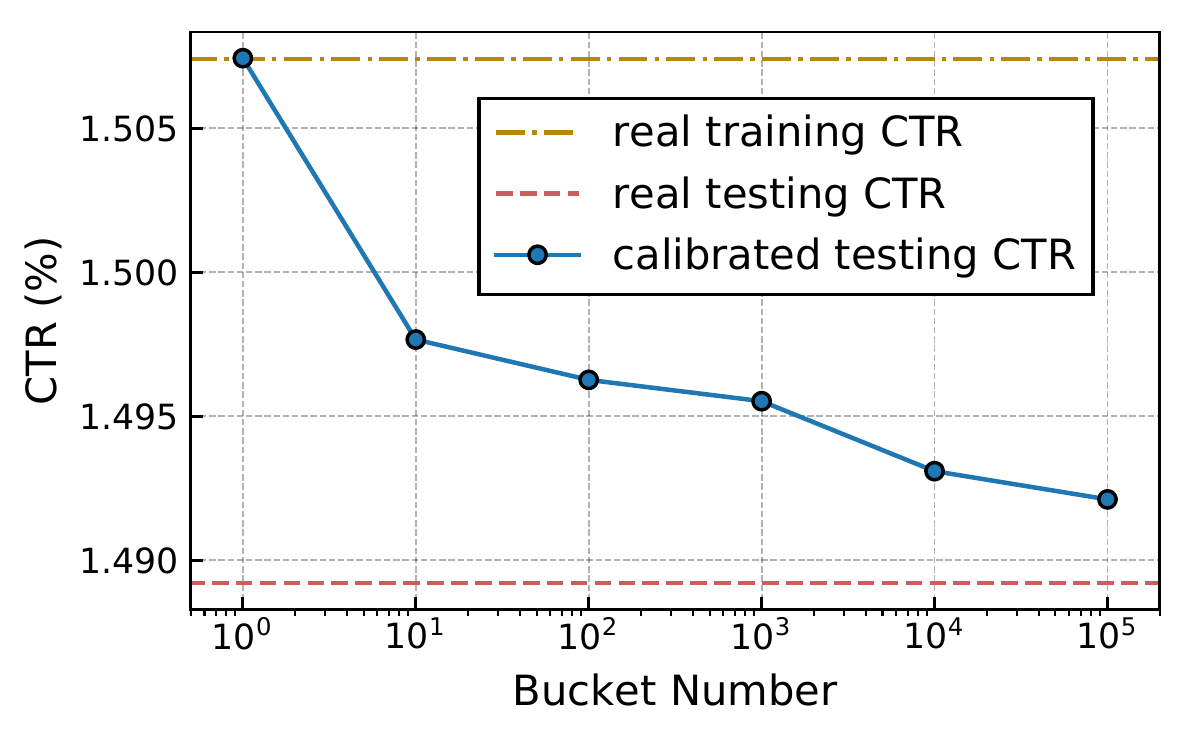}}
\hspace{-0.1in}
  \caption{Performance of the absolute CTR calibration}
  \label{cal}
\end{figure}

Experiments above indicate that the proposed models bring markedly better relative CTR values which will benefit the ranking process in CPC advertising. Considering the bidding process, we now follow subsection \ref{subsec33} to calibrate the absolute CTR values. For the In-station-Sep. dataset, Figure \ref{cal}(a) illustrates the curves of $p(v_j)$ in Eq. (\ref{pvj}) with $300$ buckets (denote as $n$) and the corresponding isotonic regression fitting using PAVA\cite{Leeuw2009Isotone}. We add a small slope $\epsilon=0.1$ on the isotonic fitting curve to make it strictly increasing. Figure \ref{cal}(b) illustrates absolute CTR calibrated results on the testing data of the In-station-Sep. dataset w.r.t. various bucket numbers. With $10^5$ buckets, the calibration error drops to as low as $0.19\%$. We also verify the performance on the out-station-Jul. and out-station-May. datasets where the error are $0.20\%$ and $0.22\%$ respectively. When calibrated CTR values of some certain categories (cid3) are required, we can directly apply the same calibration procedure specifically on these certain categories.

\subsection{In-depth Model Analysis}
\label{sec:indepth}
In this part, we dive into an in-depth model analysis and quantify the benefits of the components that build-up the proposed regularized adversarial (rGAN) sampling strategy.

\begin{figure}[H]
  \centering
\hspace{-0.185in}
  \subfigure[In-station-Sep. dataset]{\includegraphics[height=1.3in,width=1.68in]{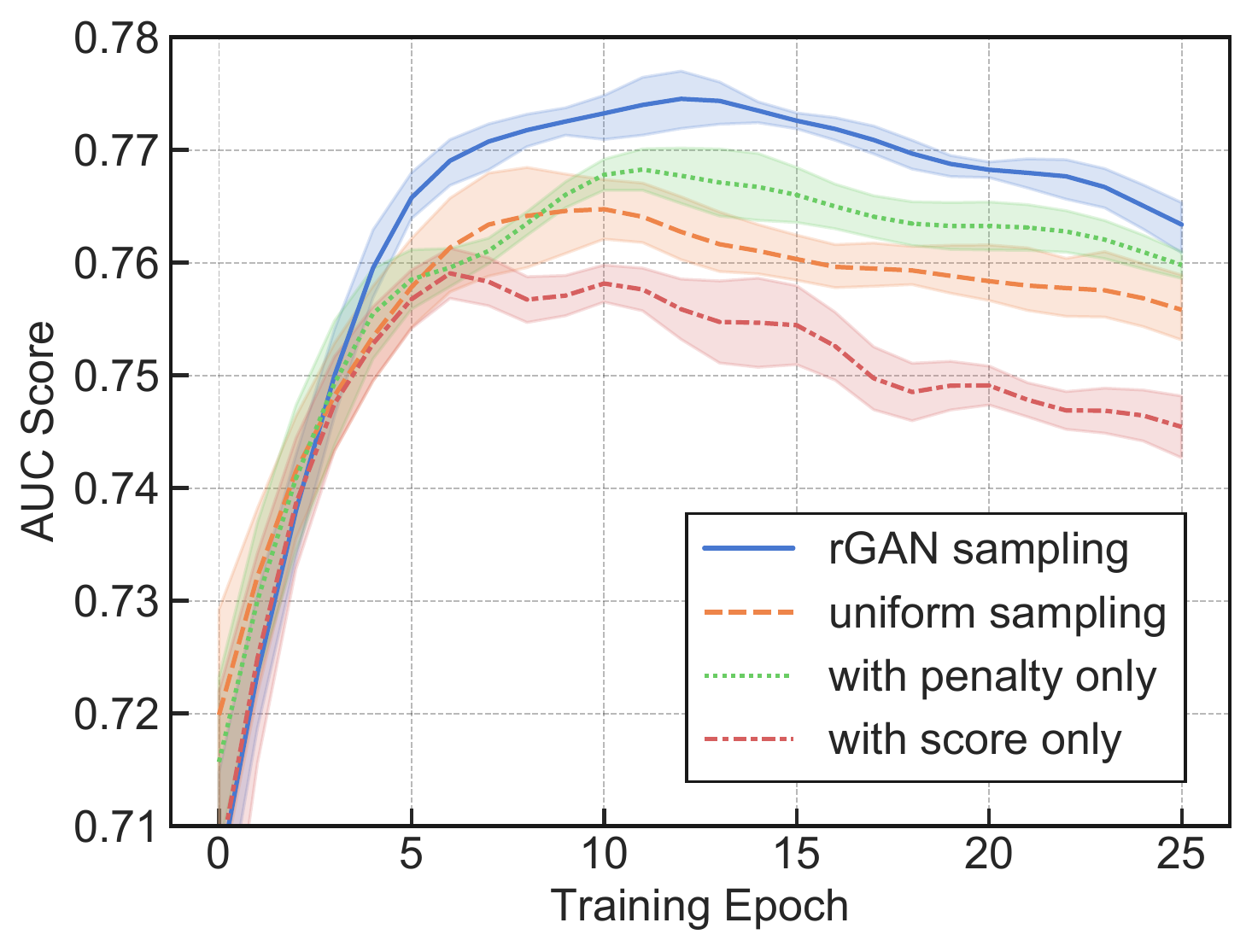}}
\hspace{-0.15in}
  \subfigure[Out-station-Jul. dataset]{\includegraphics[height=1.279in,width=1.68in]{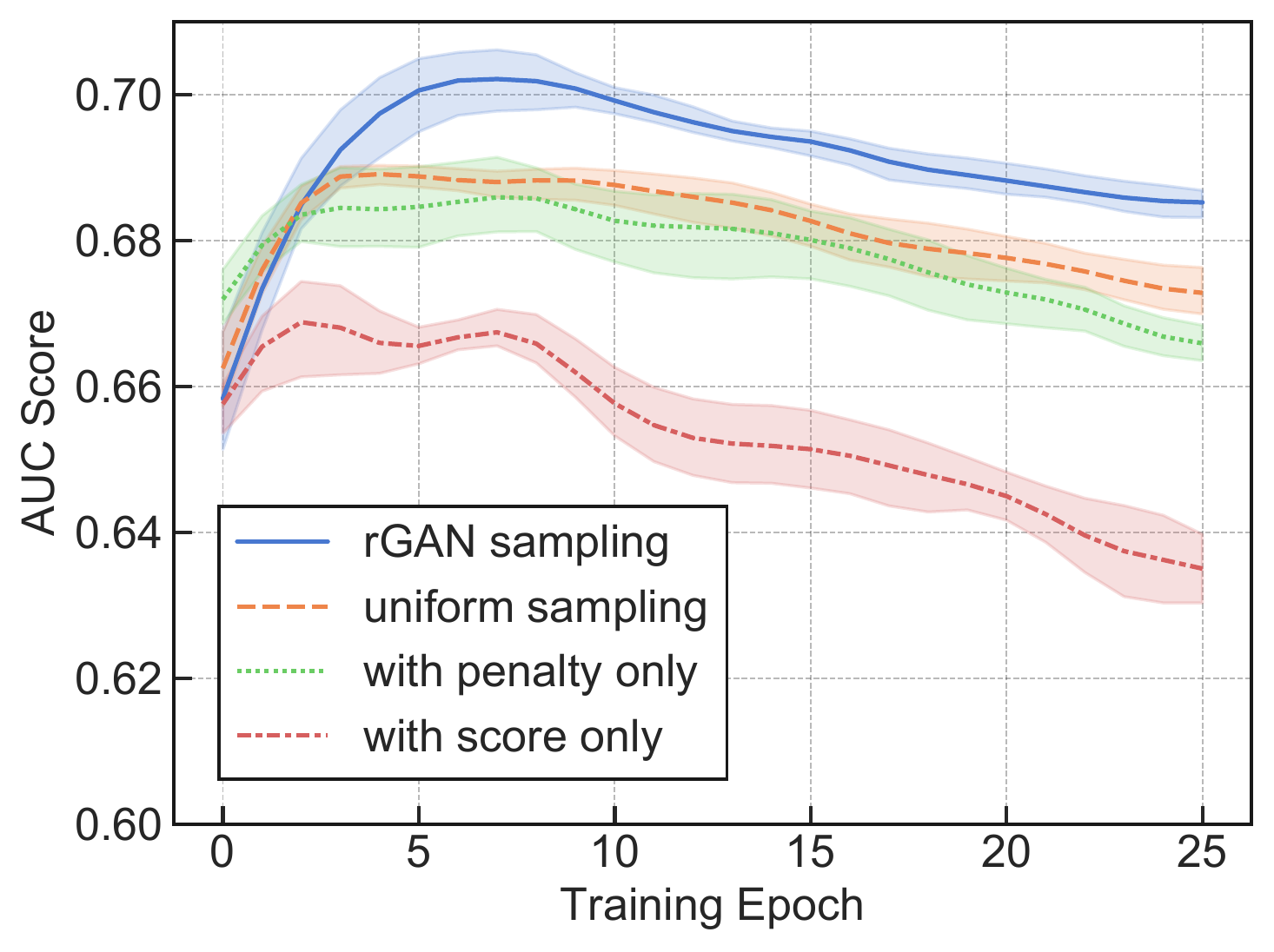}}
\hspace{-0.1in}
  \caption{Performance of various reward structures}
  \label{reg}
\end{figure}

In Eq. (\ref{reward}), the reward function of rGAN sampling contains a score term and a distance penalty term. To show the importance of the two components, we omit either one of them and conduct comparison experiments. Figure \ref{reg} illustrates that a good cooperation of the score and the penalty will improve the CTR prediction performance, and the two terms are both indispensable.

\begin{figure}[ht]
\centering
\includegraphics[scale=0.35]{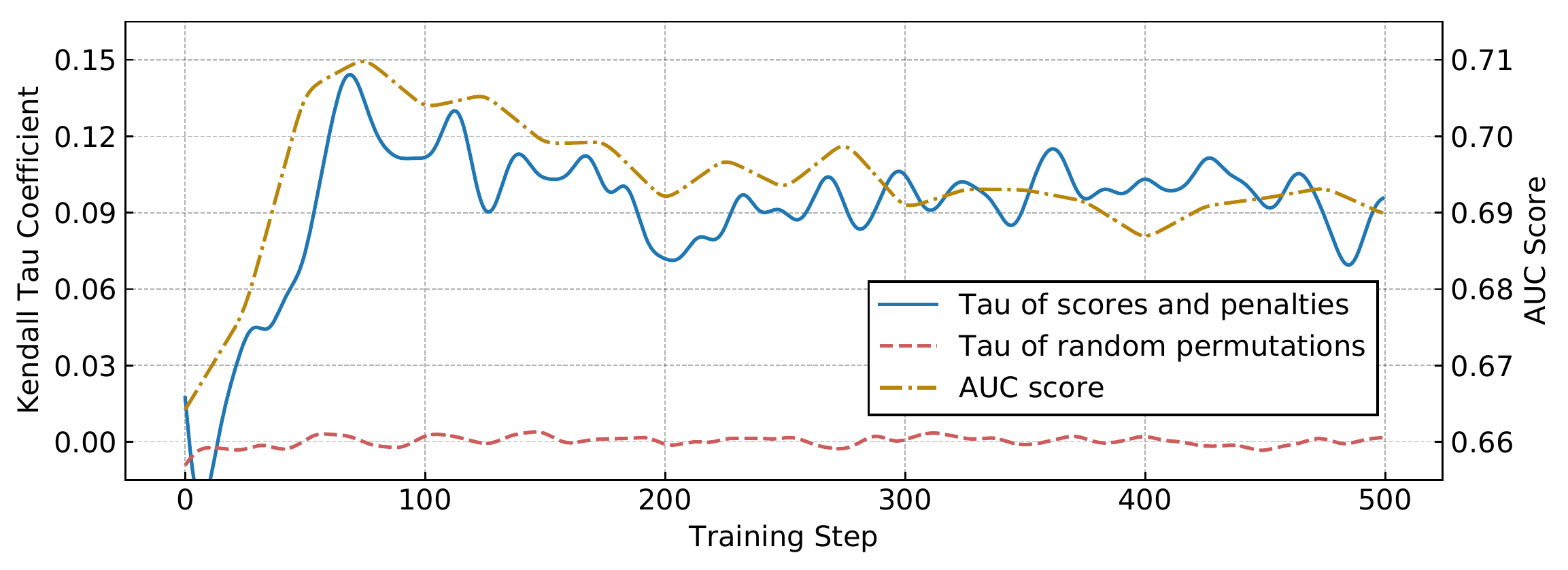}
\caption{Tau coefficient of scores and negative penalties}
\label{tau}
\end{figure}

We further explore the relation between scores and penalties during training. For each positive sample $s$, we rank the selected $C$ candidate negatives $s'_j$ according to their scores $f_D(\bm{e}_D^s(s'_j))$ and their negative penalties $-p(s,s'_j)$ respectively, where $j=1,\cdots,C$. Thus we get two permutations for $s$ both with length $C$. We apply Kendall Tau correlation coefficient\cite{tau} to measure the similarity between the permutations, where a higher Tau coefficient indicates more similarity between the permutations. We average over the Tau coefficients of the mini-batch positive samples for each training step. Figure \ref{tau} illustrates the Tau coefficient curve in out-station-Jul. dataset. We also provide Tau curve between the permutation of score and a random permutation for comparison. The corresponding AUC score curve is together plotted in the figure, which reaches its maximum where the permutations have high Tau coefficient.

We provide the sensitivity analysis of two key hyperparameters in the regularized adversarial (rGAN) sampling, the candidate size and the initial temperature. Other parameters follow the default settings described in Section \ref{details}.

\begin{figure}[h]
  \centering
\hspace{-0.185in}
  \subfigure[In-station-Sep. dataset]{\includegraphics[height=1.05in,width=1.68in]{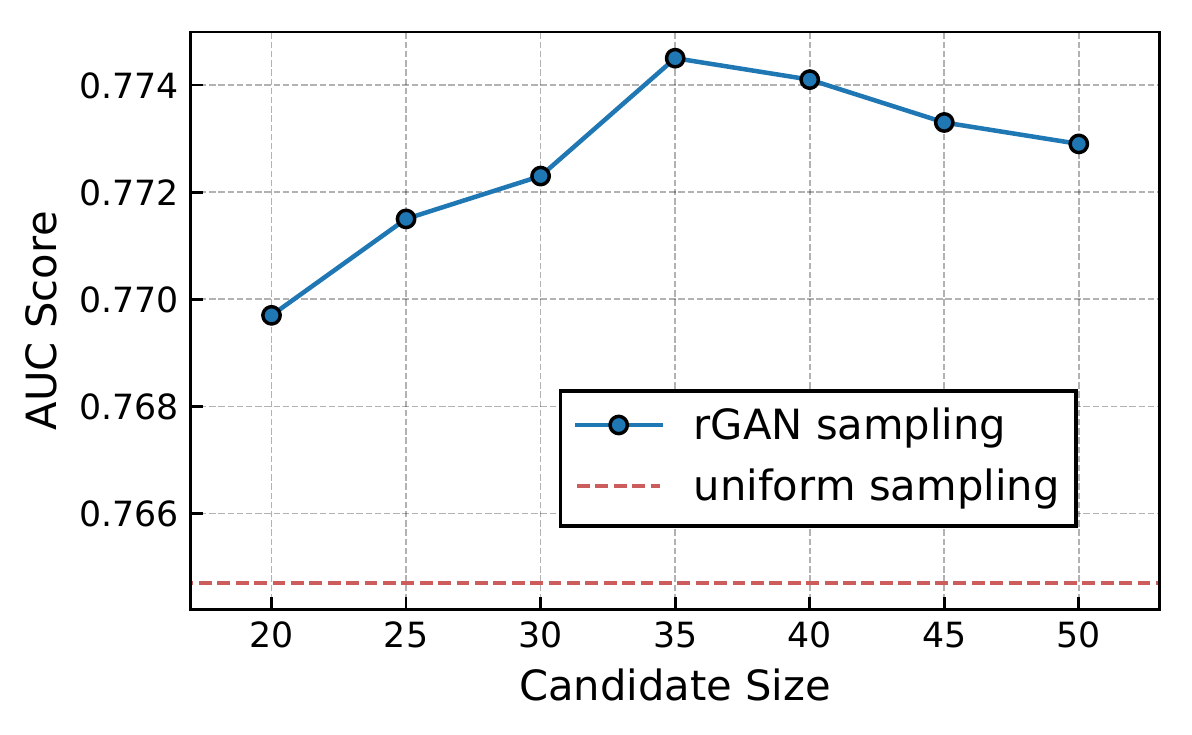}}
\hspace{-0.15in}
  \subfigure[Out-station-Jul. dataset]{\includegraphics[height=1.05in,width=1.68in]{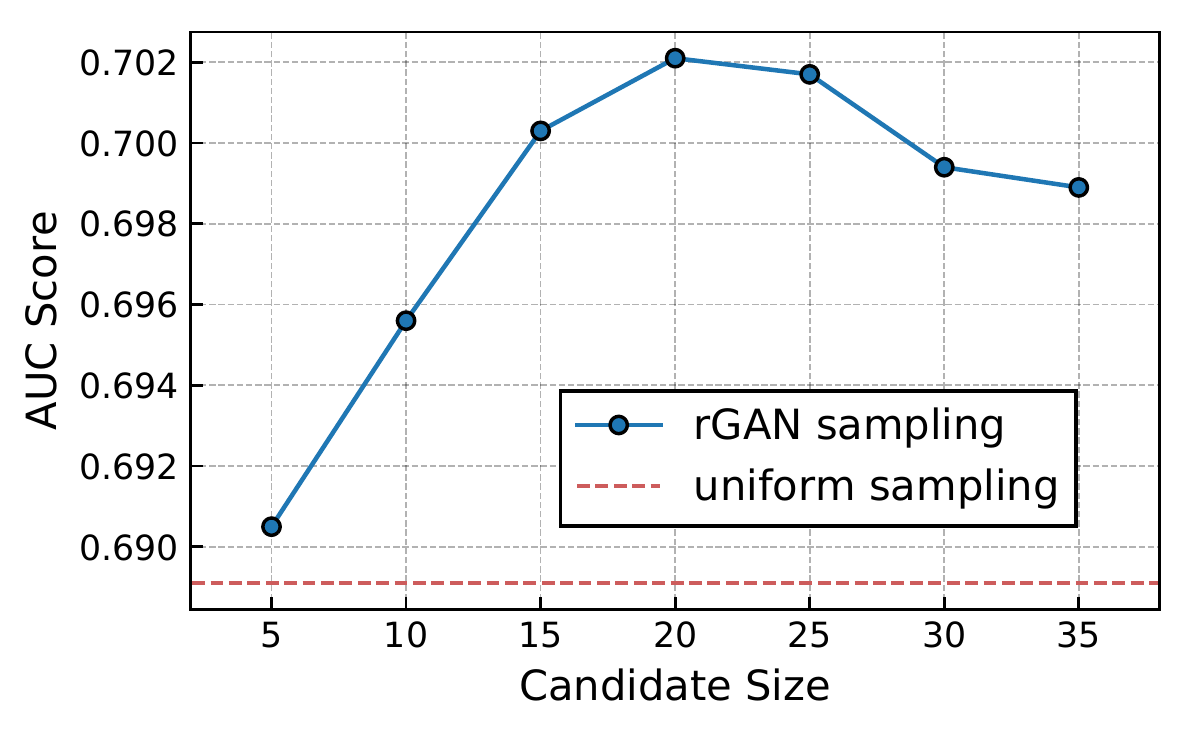}}
\hspace{-0.1in}
  \caption{Sensitivity analysis of the candidate size}
  \label{sen_cs}
\end{figure}

{\bf Impact of candidate size.} The candidate size $C$ in rGAN sampling is an essential hyperparameter. rGAN with a small candidate size will degenerate into the uniform sampling, and with a large candidate size will be vulnerable to the false negatives (as analyzed in Section \ref{sub:sampling_strategy}). We test the sensitivity of the candidate size using In-station-Sep. and Out-station-Jul. datasets, and show the average results in Figure \ref{sen_cs}. Candidate size has obvious impact on the performance and needs adjusting according to both the data amount and the noise level.

\begin{figure}[h]
  \centering
\hspace{-0.185in}
  \subfigure[In-station-Sep. dataset]{\includegraphics[height=1.05in,width=1.68in]{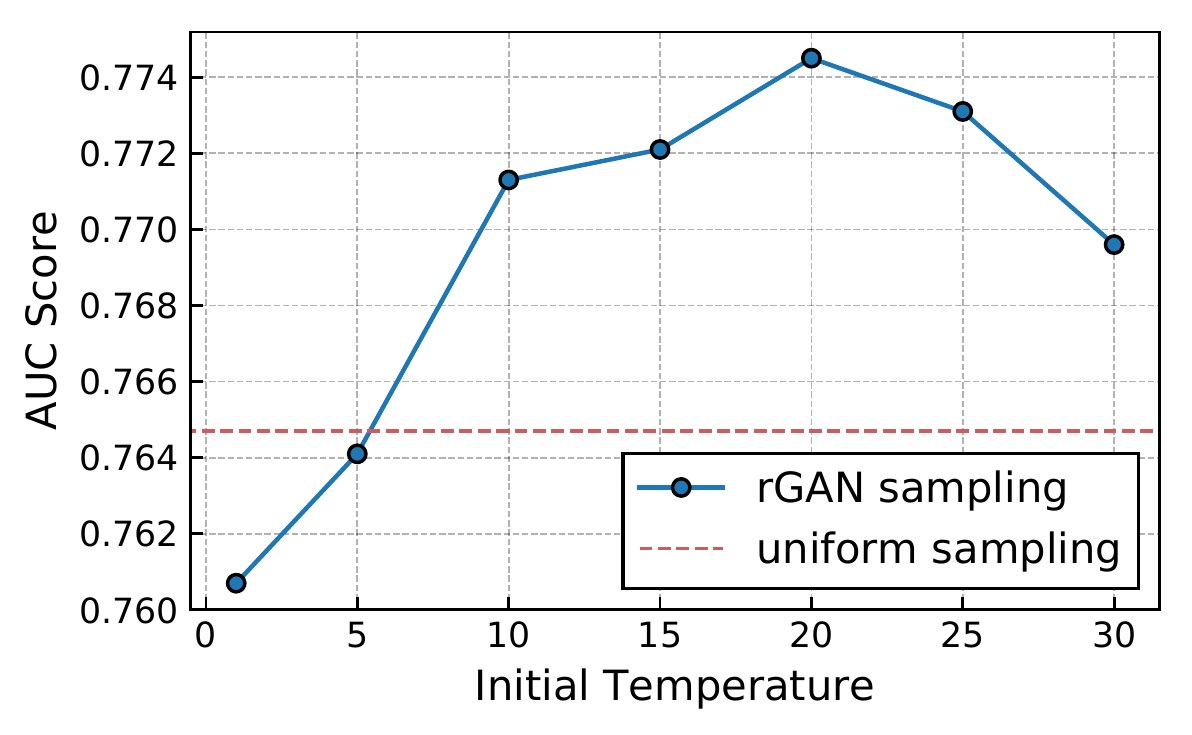}}
\hspace{-0.15in}
  \subfigure[Out-station-Jul. dataset]{\includegraphics[height=1.05in,width=1.68in]{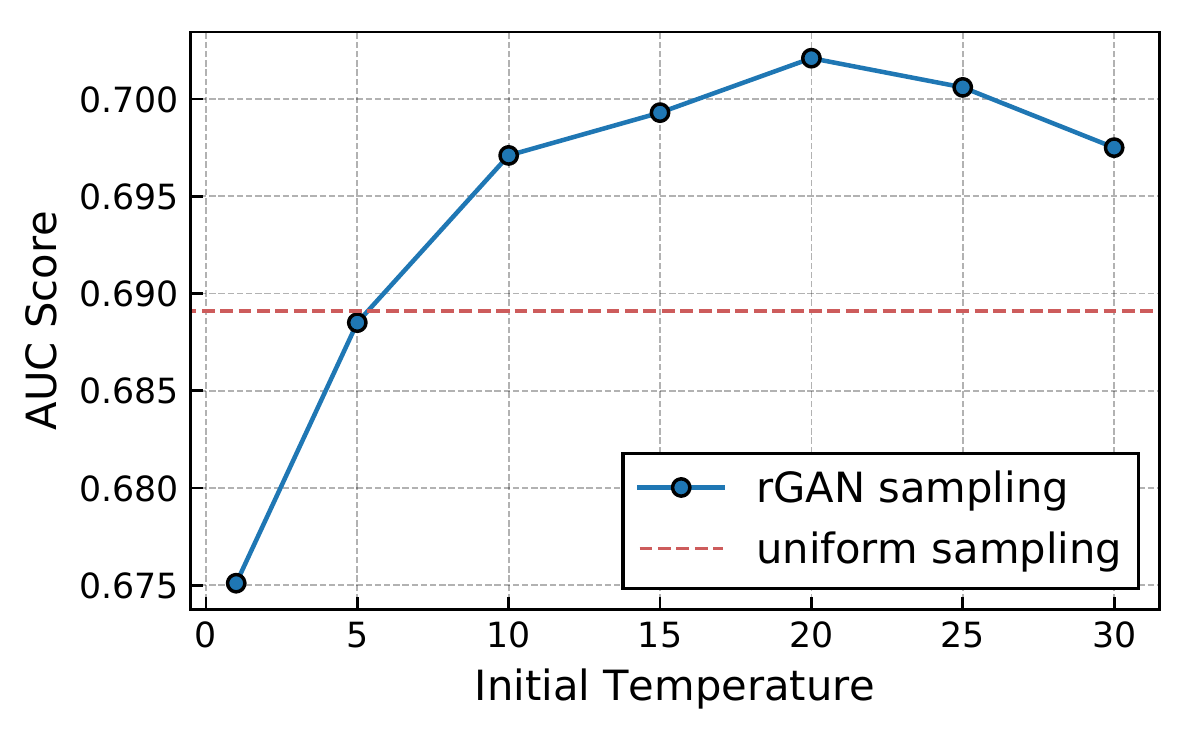}}
\hspace{-0.1in}
  \caption{Sensitivity analysis of the initial temperature}
  \label{sen_tem}
\end{figure}

{\bf Impact of initial temperature.} As analyzed in Section \ref{sub:sampling_strategy}, the temperature $T$ is used to adjust the sensitivity of Eq. (\ref{softmax}). We conduct experiments to observe the results with various initial temperatures shown in Figure \ref{sen_tem}. 

Results illustrate that the AUC score increases firstly and then decreases as temperature increases.  The two datasets both reach the best performance when $T=20$, which shows that the temperature has good generalization to datasets.

\section{Related Work}
\label{sec5}
We summarize the related works according to the deep CTR prediction models and the adversarial negative sampling methods. 

With the rise of deep learning, CTR prediction models have evolved from shallow to deep, aiming to improve the model capacities. Typical models like Wide$\&$Deep\cite{DBLP:conf/recsys/Cheng0HSCAACCIA16} which combines low-order logistic regression components and high-order features with neural networks for better representation of feature interactions. 
PNN\cite{DBLP:conf/icdm/QuCRZYWW16} imposes networks based on inner and outer products for stronger expression of cross features. DCN\cite{DBLP:conf/kdd/WangFFW17} introduces a novel cross network which is more efficient in learning feature interactions explicitly by apply feature crossing at each layer. These deep CTR works greatly enhance model capacities, while lack the deeper exploration of the users' historical behaviors. With the evolution of the recommender systems, user-item interactions are detailedly recorded. DIN\cite{DBLP:conf/kdd/ZhouZSFZMYJLG18} activates historical behaviors regarding to the target item locally with attention mechanism to capture users' relative interests. DIEN\cite{DBLP:conf/aaai/ZhouMFPBZZG19} designs an attention-based interest extractor layer to capture diverse interests from users' historical records. However, these state-of-the-art deep CTR prediction models neglect the importance of the temporal signals in users' historical records. Our proposed time-aware attention model aims to improve the performance of CTR prediction giving credit to the temporal signals, which reflect the users' periodic trends and measure the temporal influence of each historical item to the target recommended item. 

Generative adversarial nets (GANs), which are originally proposed to fit continuous data distributions\cite{DBLP:conf/nips/GoodfellowPMXWOCB14}, have been recently used for negative sampling in discrete data to promote the training efficiency. IRGAN\cite{DBLP:conf/sigir/WangYZGXWZZ17} unifies generative and discriminative models of information retrieval into a discrete GAN framework. AdvIR\cite{DBLP:conf/www/ParkC19} expands IRGAN by adding additional generated adversarial examples for joint training. \cite{DBLP:conf/acl/BoseLC18} learns by contrasting observed and fictitious samples with an adversarially learned sampler. KBGAN\cite{DBLP:conf/naacl/CaiW18} designs an adversarial framework with dual KGE components for improving knowledge graph embedding models. \cite{DBLP:conf/kdd/WangYHLWH18} further extends the idea to recommender systems, and proposes an adaptive adversarial negative sampling scheme to generate negative samples for each user, where the item in each negative sample is selected from all items not interacted with the user. A common point of these works is that, each negative sample is a combination of two components with few interactions. In many application domains such as CTR prediction tasks, however, observed negative interactions are available as well, which provide stronger negative guidance than nonpositive user-item combinations. Our regularized adversarial sampling is designed for CTR prediction tasks and unlike these common adversarial sampling strategies, it can make use of the strong information of the observed negative samples.

\section{Conclusion}
\label{sec6}
In this paper, we focus on designing a temporal embedding model and an adversarial sampling strategy that can promote the performance
of CTR prediction tasks. The proposed attention-based model is capable to represent the users' periodic behaviors and the temporal relations between historical items and target items, by considering absolute and relative temporal signals. In addition, the proposed regularized adversarial negative sampling is able to make use of the stronger guidance provided by the negative CTR samples, which is different from existing adversarial sampling methods in recommender systems. And the idea of regularization in adversarial sampling can be potentially extend to other fields. We test our models in three real-world CTR datasets. Comparison results show that the collaboration of the time-aware attention and the regularized adversarial sampling strategy outperforms the state-of-the-art CTR prediction models. 

\begin{acks}
This work is supported by the National Natural Science Foundation of China under Grants 61621136008 and 91848206. 
\end{acks}

\bibliographystyle{ACM-Reference-Format}
\bibliography{cite}

\end{document}